\pdfoutput=1
\documentclass[11pt]{article}

\usepackage[final]{acl}

\usepackage{times}
\usepackage{latexsym}

\usepackage[T1]{fontenc}
\usepackage[utf8]{inputenc}

\usepackage{microtype}

\usepackage{inconsolata}

\usepackage{graphicx}

\usepackage{caption}
\usepackage{subcaption}
\usepackage{enumitem}
\usepackage{multirow}
\usepackage{booktabs}
\usepackage{hyperref}
\newcommand{\rom}[1]{\lowercase\expandafter{\romannumeral #1\relax}}

\title{Randomly Removing 50\% of Dimensions in Text Embeddings\\has Minimal Impact on Retrieval and Classification Tasks}

\author{Sotaro Takeshita\textsuperscript{1}, Yurina Takeshita\textsuperscript{2}, Daniel Ruffinelli\textsuperscript{1}, Simone Paolo Ponzetto\textsuperscript{1} \\
  \textsuperscript{1}Data and Web Science Group, University of Mannheim, Germany \\
  \textsuperscript{2}Independent researcher \\ 
  \texttt{\{sotaro.takeshita, druffinelli, ponzetto\}@uni-mannheim.de}}

\begin{document}
\maketitle

\begin{abstract}
    In this paper, we study the surprising impact that truncating text 
    embeddings has on downstream performance.
    We consistently observe across 6 state-of-the-art text encoders and 26 
    downstream tasks, that randomly removing up to 50\% of embedding dimensions 
    results in only a minor drop in performance, less than 10\%, in retrieval 
    and classification tasks.
    Given the benefits of using smaller-sized embeddings, as well as the 
    potential insights about text encoding, we study this 
    phenomenon and find that, contrary to what is suggested in prior work, this 
    is not the result of an ineffective use of representation space.
    Instead, we find that a large number of uniformly distributed dimensions
    actually cause an increase in performance when removed.
    This would explain why, on average, removing a large number of embedding
    dimensions results in a marginal drop in performance.
    We make similar observations when truncating the embeddings used by large
    language models to make next-token predictions on generative tasks, 
    suggesting that this phenomenon is not isolated to classification or 
    retrieval tasks.
    Our code is attached to the submission.\footnote{\url{https://sotaro.io/papers/trunbed}}
\end{abstract}

\section{Introduction}
As text embeddings are used in various applications such as retrieval augmented generation~\citep{li-etal-2025-enhancing-retrieval}, question answering~\citep{karpukhin-etal-2020-dense}, or text retrieval~\citep{liu-etal-2021-improving-embedding-based}, there have been extensive research efforts not only aiming at improving their performance but also to understand them. A number of works explore \textit{what} text embeddings encode, such as using probing methods in well-controlled setups to check the information encoded by embeddings~\citep{hewitt-manning-2019-structural,kulmizev-etal-2020-neural}.
However, less has been explored on \textit{how} information is encoded. Existing works in this direction often assess isotropy (or anisotropy) in text embeddings, that is, whether text embeddings are scattered (or concentrated) in representation space \citep{ait-saada_anisotropy_2023,godey_anisotropy_2024}. While these works provide insights into geometric properties of text embeddings, they often focus less on the impact that these properties have on downstream tasks.

\begin{figure}
    \centering
    \begin{subfigure}{\columnwidth}
        \centering
        \includegraphics[width=0.9\textwidth]{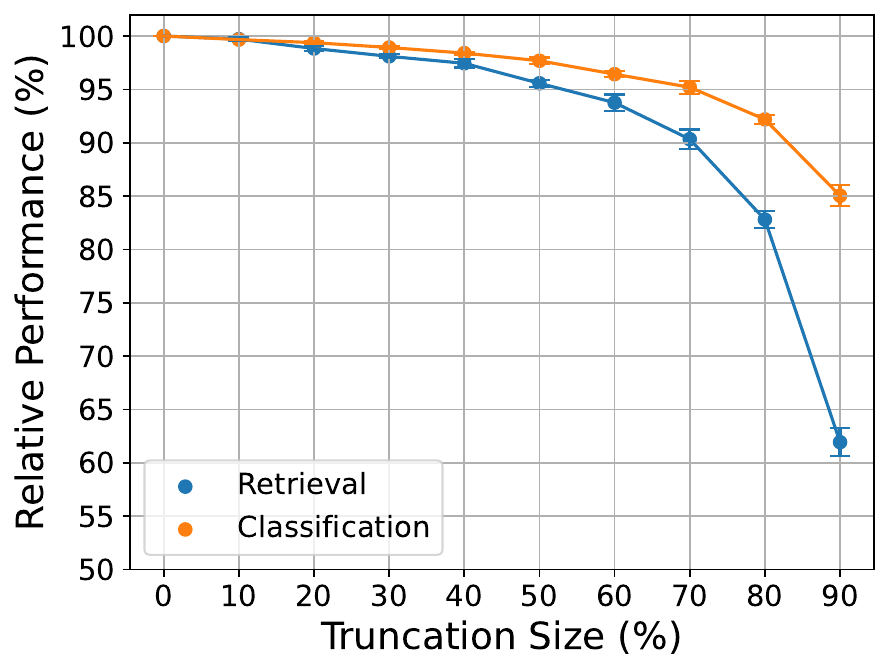}
    \end{subfigure}
    \begin{subfigure}{\columnwidth}
        \centering
        \includegraphics[width=\textwidth]{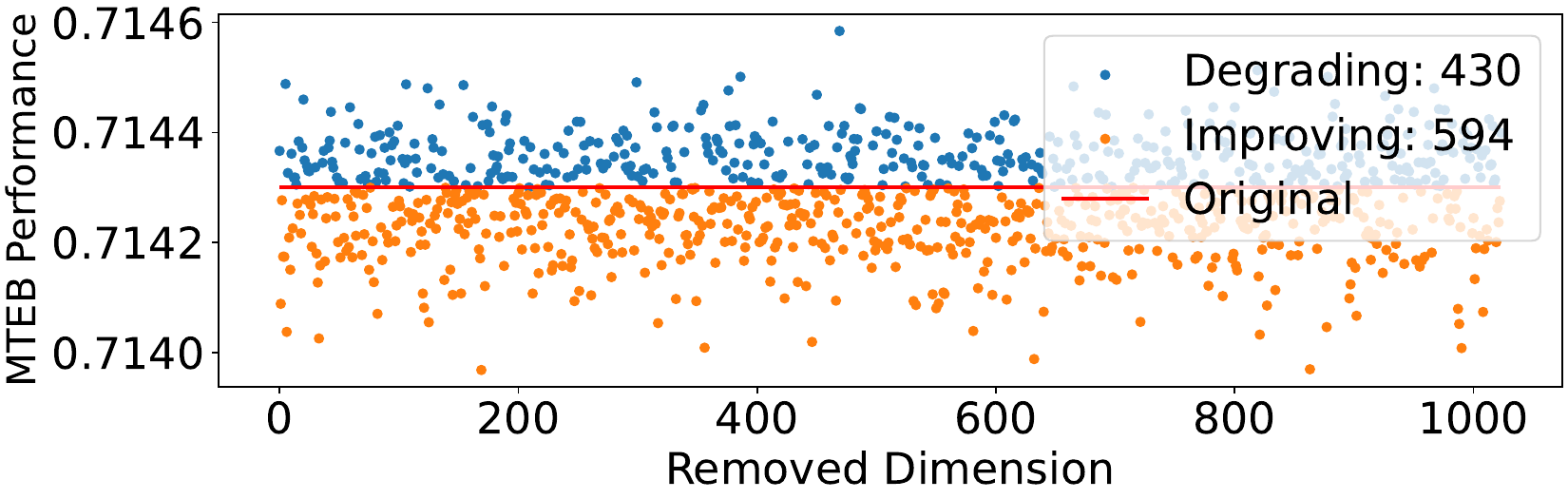}
    \end{subfigure}
    \caption{(top) Regardless of the selection, removing 50\% of embedding dimensions results in less than 5\% performance drop. (bottom) This seems related to is due to many dimensions that lower performance (depicted in blue).}
    \label{fig:first-page-figure}
\end{figure}

To fill this gap, we look at how text embeddings use the representation space \textit{through the lens of their impact on downstream task performance.}
In our first experiment, we examine how well text encoders use the embedding space by measuring performance when removing K\% of embedding dimensions. Through extensive testing with 6 text encoders, including a large language model (LLM), and 26 embedding-based tasks (e.g., passage retrieval and intent classification), we find that, surprisingly, the resulting embeddings still achieve comparable performance to the original embeddings even when more than half of the dimensions are removed. Regardless of the selection of removed dimensions, reduced embeddings can retain 95\% and 90\% of the original performance, in classification and retrieval tasks, respectively (Fig.~\ref{fig:first-page-figure}: top).
By measuring Spearman correlation, we find that indeed, truncation seems to preserve the properties of the space well enough that the rankings used for retrieval and classification are mostly unaffected.
These results suggest an inefficient use of the representation space by text encoders.

To study whether embeddings use the representation space effectively, we integrate three well-studied concepts into our study. Specifically, we test if the cause is (\rom{1}) embeddings gathering in a narrow cone in representation space \citep{xiao-etal-2023-isotropy}, (\rom{2}) redundancy in dimensions \citep{jing_understanding_2021}, or (\rom{3}) a few outlier dimensions that determine performance \citep{kovaleva_bert_2021}. While all of these properties are present in every models, we do not find strong relations to our initial observation, calling for a new perspective.

To this end, we take inspiration from input attribution methods \citep{sundararajan_axiomatic_2017,bastings_will_2022}, and analyze the contribution of each embedding dimension on downstream performance.
We find that every model contains many dimensions that negatively impact performance, dubbed degrading dimensions.
For instance, we identify 430 degrading dimensions (out of 1024) in E5-large~\citep{wang_text_2022}~(Fig.~\ref{fig:first-page-figure}: bottom).
The degrading dimensions are uniformly distributed across embedding features.
This suggests a possible reason for our initial observation, that is, when we remove dimensions randomly, both the positively and negatively contributing features are removed, resulting in a marginal performance drop.
Indeed, when removing only the degrading dimensions, the performance drop is much slower compared to removing random dimensions, or the performance improves from the original embeddings.
We also find that a significant number of degrading dimensions are shared across tasks, indicating the potential for further improvements in text encoders.

While our focus is on embeddings produced by text encoders, we also find similar results when truncating the embeddings used by LLMs for next-token prediction in text generation tasks, i.e., tasks where these truncated embeddings are repeatedly used. However, the results in this case are more task-dependent, as we find that in some tasks, performance is severely degraded.

Our contributions are the following:
\begin{itemize}[itemsep=0mm,leftmargin=6mm]
    \item We consistently find across several models and downstream tasks, that
          text embeddings retain more than 90\% of their original performance 
          even after randomly removing 50\% of their dimensions. We make similar
          but less consistent findings about the embeddings used by causal language 
          models.
    \item We identify that the representations obtained from state-of-the-art
          text embedders can contain a significant amount of dimensions that
          have a negative impact on many downstream tasks, but more research is
          needed to understand their role in text representations and 
          to possibly improve existing text encoders.
\end{itemize}

\begin{figure*}[t]
    \centering
    \begin{subfigure}{0.51\columnwidth}
        \centering
        \includegraphics[width=\textwidth]{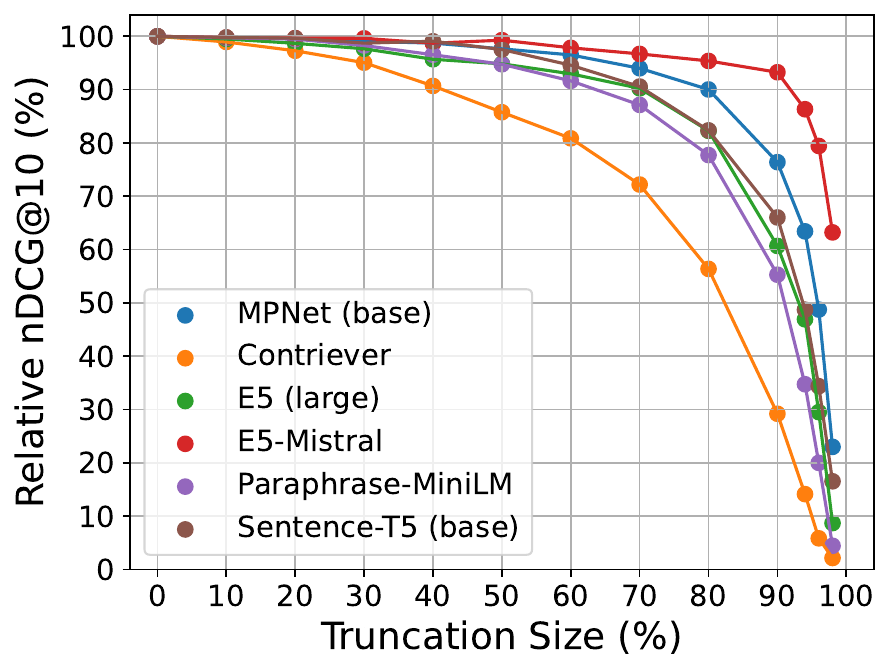}
        \caption{Last K\%, BEIR}
        \label{fig:k-truncation-last-beir}
    \end{subfigure}
    % \hfill
    \begin{subfigure}{0.51\columnwidth}
        \centering
        \includegraphics[width=\textwidth]{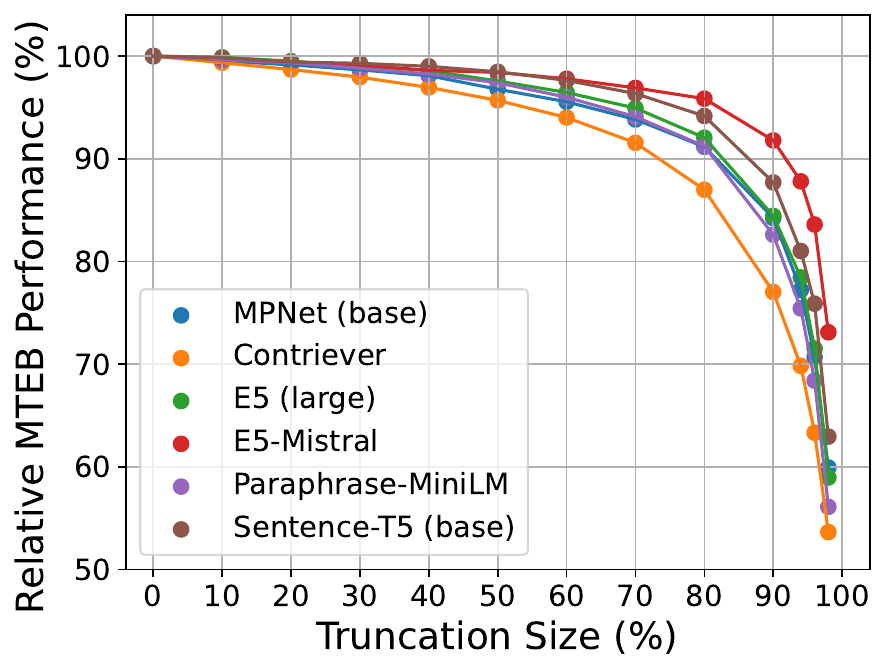}
        \caption{Last K\%, MTEB}
        \label{fig:k-truncation-last-mteb}
    \end{subfigure}
    % \hfill
    \begin{subfigure}{0.51\columnwidth}
        \centering
        \includegraphics[width=\textwidth]{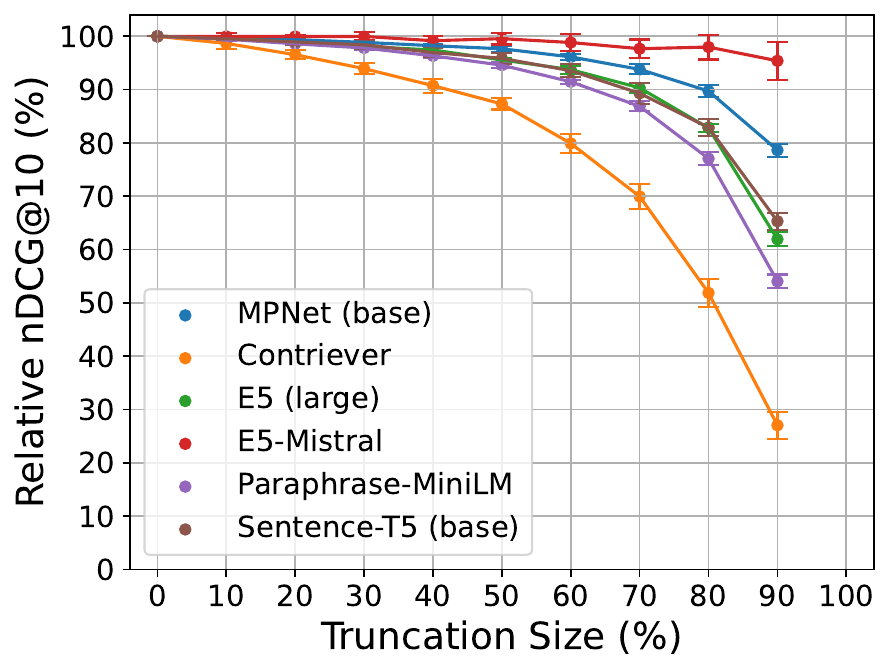}
        \caption{Random K\%, BEIR}
        \label{fig:k-truncation-random-beir}
    \end{subfigure}
    % \hfill
    \begin{subfigure}{0.51\columnwidth}
        \centering
        \includegraphics[width=\textwidth]{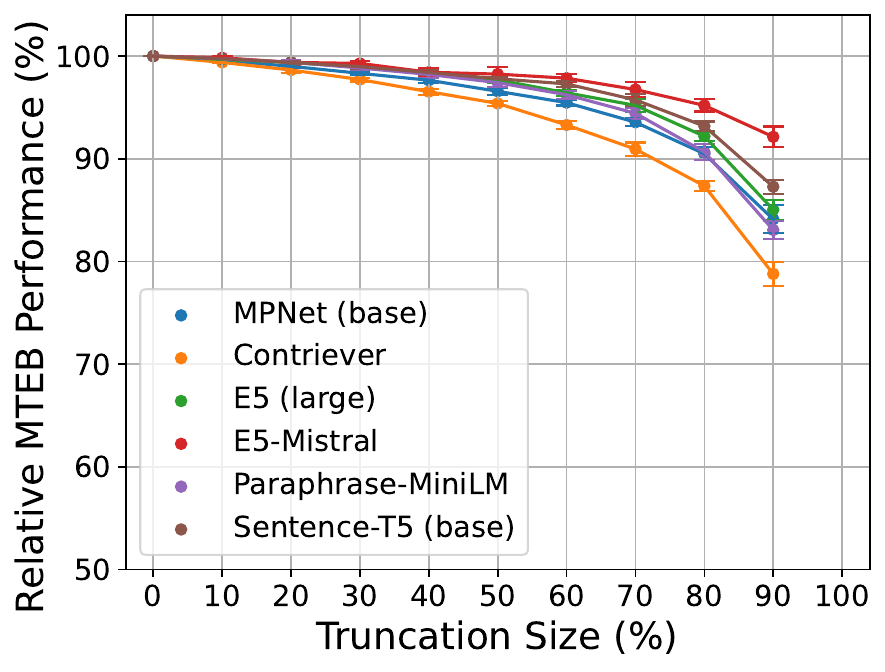}
        \caption{Random K\%, MTEB}
        \label{fig:k-truncation-random-mteb}
    \end{subfigure}
    \caption{Relative performance when (a, b) last and (c, d) random K\%  of dimensions are removed. Error bars in (c, d) are drawn from the results of ten different random removals. The results per dataset are shown in Fig. \ref{fig:k-truncation-last-beir-datasets}, \ref{fig:k-truncation-last-mteb-datasets}.}
    \label{fig:k-truncation}
\end{figure*}
\section{Impact of Embedding Truncation }
\label{sec:k-removal}
In this section, we study how well text embeddings use their dimensions by considering a simple hypothesis: if text encoders are effectively using embedding dimensions, removing some features in an arbitrary manner should have a noticeably negative effect on performance. To test this hypothesis, we look at the impact that different methods of embedding truncation have on downstream performance.

\subsection{Experimental Settings}

\paragraph{Models.}
We consider 6 state-of-the-art models, including an LLM-based model, with various sizes and training configurations. All models are contrastively trained after pre-training (see Table \ref{tab:model-list} for a list of models).

\paragraph{Tasks.}
For downstream task evaluation, we take 14 retrieval and 12 classification datasets from BEIR \citep{thakur_beir_2021}\footnote{Due to its high computational demand, for the E5-Mistral model, we use \href{https://huggingface.co/collections/sentence-transformers/nanobeir-with-bm25-rankings-67bdcbc629f007c15bf358d8}{NanoBEIR}, which is a subset of the BEIR.} and MTEB \citep{muennighoff_mteb_2023} benchmarks (see Table \ref{tab:datasets} for a list of datasets).

\paragraph{Truncation methods.}
We evaluate with two different truncation approaches: (\rom{1}) last K\% truncation: we simply remove the last K\% of dimensions from embeddings, and (\rom{2}) random K\% truncation: we uniformly sample K\% of the features to be removed. We repeat this process ten different times and report standard deviation in Fig.~\ref{fig:k-truncation-random-beir} and~\ref{fig:k-truncation-random-mteb}.

\subsection{Results and Discussion}
\paragraph{Last K\% truncation.}
\label{sec:last-k-removal}
Fig.~\ref{fig:k-truncation-last-beir} and~\ref{fig:k-truncation-last-mteb} show the results on each benchmark. 
Relative performance only drops to below 80\% when 80\% of the dimensions are removed for five out of six models. In an extreme case, E5-Mistral retains 90\% of its original performance even when 90\% of its dimensions are removed. This suggests that most of the features in these text embeddings do not have a big impact on downstream performance. We further compare the two rankings of retrieved documents produced by original embeddings and truncated embeddings, and indeed see that Spearman's rank-order correlation remains high even after truncating 50\%, higher than 0.8 for all models (details in Fig.~\ref{fig:rank_analysis_nanobeir}).

\begin{figure*}[t]
    \centering
    \includegraphics[width=1\textwidth]{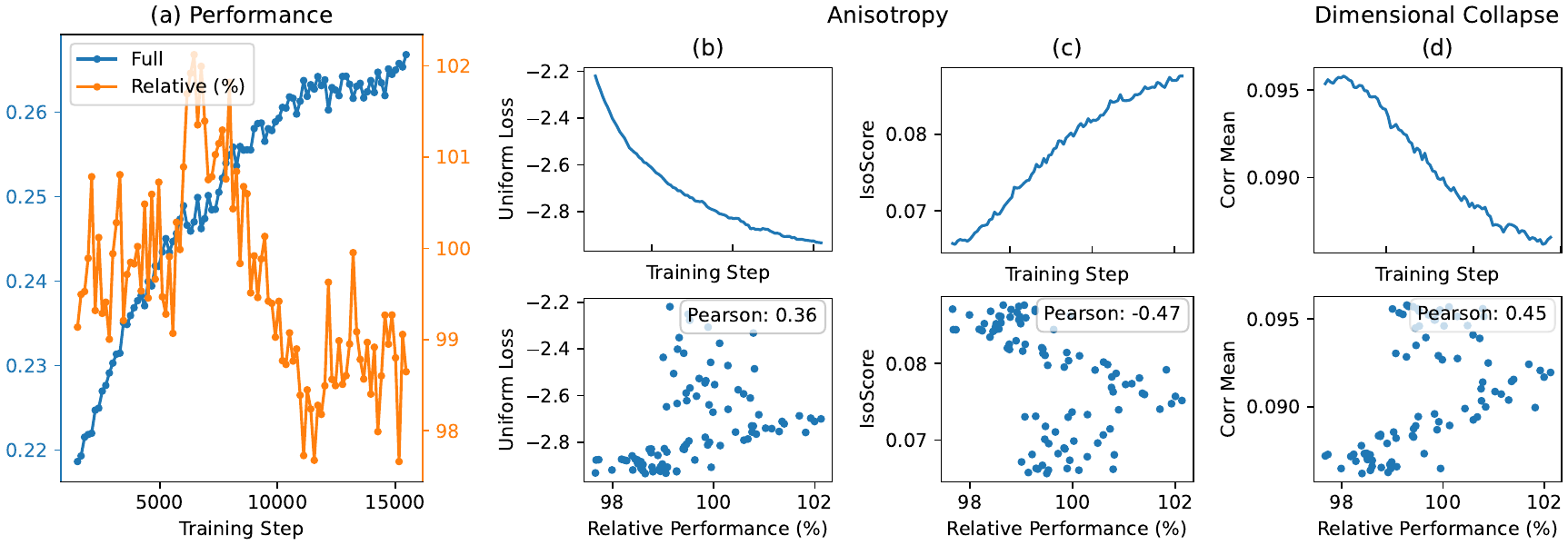}
    \caption{
    As a result of contrastive learning for T5, downstream task performance increases (a: Full), and the use of embedding space measured through Uniform Loss ($\downarrow$) and IsoScore ($\uparrow$) for anisotropy (b, c: top) and Corr Mean ($\downarrow$) for dimensional collapse (d: top) also improves.
    However, the relative performance does not change over the training (a: Relative), therefore, there is no strong correlation between relative performance and representation quality measures (b, c, d: bottom).
    }
    \label{fig:existing-theories-t5}
\end{figure*}

\paragraph{Random K\% truncation.}
\label{sec:random-k-removal}
Fig. \ref{fig:k-truncation-random-beir} and \ref{fig:k-truncation-random-mteb} show a very similar pattern in performance reduction curves compared to our previous experiments, and the standard deviations of relative performance between ten random runs are small. This means that regardless of the selection of removed dimensions, models are able to retain most of their downstream performance. In the next section, we look at existing theories in the literature that may explain this phenomenon.
\section{Effective Use of Representation Space}
\label{sec:existing-theories}
In this section, we investigate why text embeddings can be significantly reduced 
in size without much performance loss.
We do this from the perspective of three different concepts from prior 
works that explore how embeddings (in)effectively use the representation 
space: anisotropy, dimensional collapse, and outlier dimensions.

\subsection{Anisotropy in Embeddings}
\label{sec:anisotropy}
A number of existing works report that neural network-based encoders, not only for texts \citep{rudman_isoscore_2022,hammerl_exploring_2023,godey_anisotropy_2024} but also for images \citep{wang_understanding_2020}, tend to produce anisotropic embeddings, meaning that the encoders map different input data points into a narrow cone without fully exploiting the representation space. The two main characteristics of anisotropic embeddings are (\rom{1}) distorted variance in values taken by different dimensions and (\rom{2}) high correlations between different dimensions (i.e., features) \citep{rudman_stable_2023}. As this property may explain our earlier observations, we explore if anisotropy in text embeddings can be a predictor of performance with truncated embeddings.

\paragraph{Experimental setup.}
Prior work has shown that contrastively
training models makes their embeddings less anisotropic, which in turn improves downstream
performance~\citep{ni_sentencet5_2022,chen_ditto_2023}.
So, to obtain embeddings with different levels of anisotropy, we
contrastively train two pre-trained models, BERT~\citep{devlin_bert_2019} and
T5~\citep{raffel_exploring_2020}, storing intermediate checkpoints along the way.
The expectation is that, as anisotropy decreases with more training, performance
should increase for full-size embeddings, but decrease for truncated embeddings,
as less anisotropy means the model is making more effective use of the representation space.
We train the models with a mixture of SNLI~\citep{bowman_large_2015} and
MultiNLI~\citep{williams_broadcoverage_2018} as training and validation data~\citep{reimers_sentencebert_2019,gao_simcse_2021},
and use InfoNCE as a loss function~\citep{oord_representation_2019}.
The training is terminated upon convergence of the validation loss.
To measure anisotropy, we use two common metrics: uniform
loss~\citep{wang_understanding_2020,ni_sentencet5_2022} and
IsoScore~\citep{rudman_isoscore_2022}, the former decreases and the latter increases its values as the target embeddings become less anisotropic.
We apply the last K\% truncation to obtain reduced embeddings, and use 13 datasets from NanoBEIR to evaluate the final performance.

\paragraph{Results.}
Fig.~\hyperref[fig:existing-theories-t5]{3a} shows how the full-sized embeddings improve performance during the training together with the decrease in anisotropy as shown in Fig.~\hyperref[fig:existing-theories-t5]{3b, c} (top) for T5 (result for BERT is shown in Fig.~\ref{fig:existing-theories-bert}).
As expected, more training results in increased downstream performance and decreased anisotropy.
However, the relative performance achieved by half-sized embeddings does not change over training steps.
Fig.~\hyperref[fig:existing-theories-t5]{3b, c} (bottom) shows the relation between the relative performance and two anisotropy measures.
We do not see correlating patterns because even though different checkpoints have different degrees of anisotropy, their relative performance is quite stable.
We also compute the Pearson correlation between relative performance and the two anisotropy metrics, but do not observe any strong correlation (0.36 for uniform loss, -0.47 for IsoScore).
These results indicate that even when models make better use of the
representation space, as measured by anisotropy, truncated embeddings still
result in marginal performance drops.

\subsection{Dimensional Collapse}
\label{sec:dimensional-collapse}
Existing works, especially in the computer vision community, report that the representations produced by neural network-based encoders have certain dimensions collapsed, and use only a lower-dimensional subspace \citep{huang_ldreg_2023,hua_feature_2021,he_exploring_2022}. While this is conceptually similar to anisotropy, dimensional collapse focuses on the correlation between dimensions.
In this section, we explore the relation of dimensional collapse in text embeddings to the high relative performance we observe, as the embeddings with highly correlated dimensions can be robust to feature removal.

\paragraph{Experimental setup.}
We follow~\citet{hua_feature_2021} and use the mean of the correlation coefficient between dimensions, computed over embeddings obtained by encoding 10K English paragraphs from Wikipedia\footnote{\href{https://huggingface.co/datasets/sentence-transformers/simple-wiki}{sentence-transformers/simple-wiki}}.
Same as our experiments on anisotropy in \S \ref{sec:anisotropy}, we use the contrastively trained checkpoints of two model families and NanoBEIR evaluation.
We hypothesize that the impact of dimension truncation can remain low for the models where the dimensions are highly correlating.

\paragraph{Results.}
Fig.~\hyperref[fig:existing-theories-t5]{3d} (top) shows that the contrastive training reduces correlations between dimensions for T5 (result for BERT is shown in Fig.~\ref {fig:existing-theories-bert} in the Appendix). As shown in existing works \citep{jing_understanding_2021,he_exploring_2022}, downstream task performance improves as the correlations weaken. However, as shown in Fig.~\hyperref [fig:existing-theories-t5]{3d} (bottom), we do not observe any relation between feature correlations and how performance drops after truncating the last 50\% of dimensions; the Pearson correlation between them is 0.45. Given these results, we conclude that dimensional collapse, i.e.\ correlation between dimensions, does not explain the success of truncated embeddings.

\begin{table}[t]
    \centering
    \small
    \setlength{\tabcolsep}{5.0pt}
    \begin{tabular}{lrrr}
    \toprule
     & \multicolumn{1}{r}{\textbf{\# of Outliers}} & \multicolumn{1}{r}{\textbf{Outlier}} & \multicolumn{1}{r}{\textbf{Non-outlier}} \\
    \midrule
    MPNet (base) & 5 & 0.576 & 0.576 $\pm$ 9e-04 \\
    Contriever & 3 & 0.525 & 0.524 $\pm$ 1e-03 \\
    E5 (large) & 1 & 0.577 & 0.577 $\pm$ 4e-04 \\
    E5-Mistral & 32 & 0.651 & 0.626 $\pm$ 6e-04 \\
    Para-MiniLM & 2 & 0.483 & 0.483 $\pm$ 6e-04 \\
    ST5 (base) & 2 & 0.489 & 0.489 $\pm$ 9e-04 \\
    \bottomrule
    \end{tabular}
    \caption{
    Effect of removing outlier dimensions on downstream task performance.
    As a comparison, we also remove the same number of ten different non-outlier dimensions and report the average and standard deviation of achieved performance.
    }
    \label{tab:without-outlier}
\end{table}
\subsection{Outlier Dimensions in Embeddings}
\label{sec:outlier-dimensions}
Several papers report that there are a few dimensions in the weights of pre-trained language models (PLMs) that take abnormally high or low values, known as outlier dimensions \citep{kovaleva_bert_2021,puccetti_outlier_2022,hammerl_exploring_2023}. PLM weights are often redundant and can be removed without much performance loss~\citep{michel_are_2019,bian_attention_2021}; however, \citet{kovaleva_bert_2021} show that removing such outlier dimensions from models, even though there are only a few, can massively reduce performance. A follow-up work by \citet{hammerl_exploring_2023} reports that a similar trend exists in multilingual models as well. While they study outlier dimensions in model weights, in this paper, we extend this concept to identify outlier dimensions in text embeddings and explore their interactions with our interest, the high relative performance.

\paragraph{Experimental setup.} First, we aim to identify outlier dimensions within text embeddings produced by text encoders. To this end, we examine the embedding obtained by averaging the embeddings of all the query texts from the NanoBEIR datasets. We follow the definition of outlier dimensions used by \citet{kovaleva_bert_2021}, that is, the dimensions that deviate more than $3\sigma$ from the standard deviation of all values in the average embedding. After identifying them, we assess their effects on downstream tasks by comparing the performance achieved by: (\rom{1}) the embeddings without outlier dimensions, and (\rom{2}) the embeddings without non-outlier dimensions, the same number as outlier dimensions. The second configuration is our control trial. We experiment by removing ten different sets of non-outlier dimensions.

\paragraph{Results.} The number of outlier dimensions for each model and their effect on performance are shown in Table \ref{tab:without-outlier}. The number of outlier dimensions changes with different models; however, it remains low in all cases. This is similar to the outliers within weights reported by \citet{kovaleva_bert_2021}. The highest proportion of outlier dimensions to the full embedding is observed with E5-Mistral, that is 0.8\% (32 out of 4096 dimensions).
The figure also shows that, for five out of six models, the effect of removing outlier dimensions is not beyond the ones where we remove non-outlier dimensions, indicating that, outlier dimensions do not play a critical role.
The exception is E5-Mistral, where removing outlier dimensions results in a slight increase in performance
with regard to non-outlier removal counterparts, but the gap to the average performance non-outlier removal runs is too small, 0.025 in nDCG@10, to explain our initial observation.
These observations are strong indications that the outlier dimensions are unlikely to be the reasons for the low drop rate, which is our interest in this paper.
\begin{figure*}[!t]
     \centering
     \begin{subfigure}[b]{\textwidth}
         \centering
         \includegraphics[width=\textwidth]{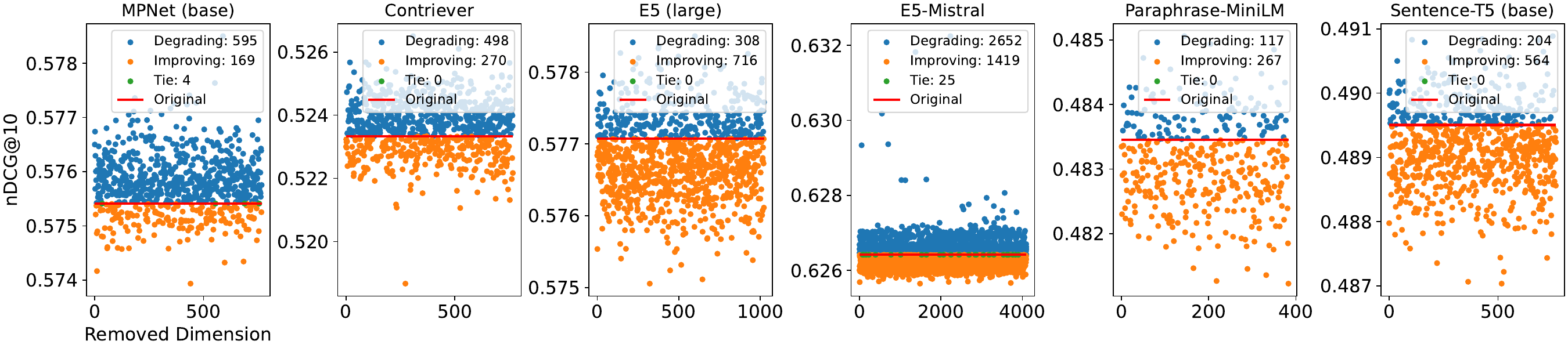}
        \caption{NanoBEIR}
        \label{fig:one-dim-drop-beir}
     \end{subfigure}
     \hfill
     \begin{subfigure}[b]{\textwidth}
         \centering
         \includegraphics[width=\textwidth]{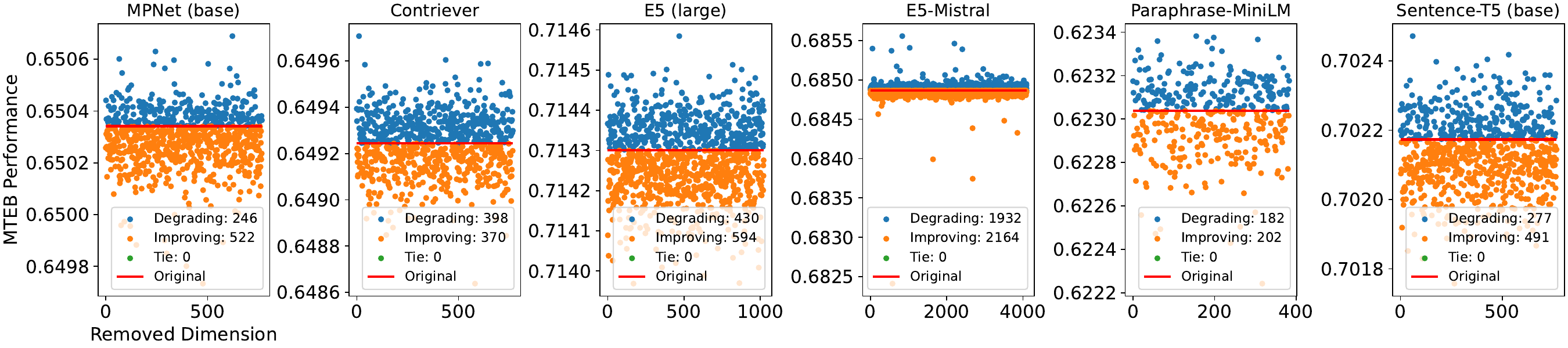}
         \caption{MTEB}
         \label{fig:one-dim-drop-mteb}
     \end{subfigure}
     \caption{
        Average performance on all datasets in the NanoBEIR and MTEB 
        benchmarks after removing each dimension in the input embeddings.
        The red horizontal line indicates the performance achieved by the 
        original embedding, and each point is the performance without the 
        corresponding dimension. Blue points indicate that they are negatively impacting the performance as they are above the red line.
     }
     \label{fig:one-dim-drop}
\end{figure*}
\section{Dimension Attribution Analysis}
\label{sec:dimension-by-dimension}
Our previous experiments suggest that truncated embeddings perform well on downstream tasks, even when they seem to make good use of the representation space through existing concepts, e.g.\ when 
embeddings are better spread across the representation space,
or when there is less correlation between features,
or when outlying dimensions are considered.
Since this implies that whatever dimensions are left after truncation are still
useful, in this section, we study the impact that each dimension has on 
downstream performance.

\paragraph{Method.} We take inspiration from existing works on input attribution, a family of methods that rely on perturbing inputs to determine feature importance to explain model predictions~\citep{sattarzadeh_explaining_2021,wu_adkd_2023}. Specifically, we repeatedly evaluate performance by disabling one dimension at a time. This enables us to measure the contribution of each dimension to the downstream task performance in isolation. Zeroing model weights is the common way to analyze their role in model behavior~\citep{serrano_attention_2019,zhang_same_2024}. However, as we focus on dimensions only in embeddings, we simply remove the target dimension from the embeddings. This is a preferred approach because, as each dimension can take a wide variety of values, zeroing values may have varying impacts on final performance.

\paragraph{Experiments setup.}
We take 13 retrieval and 12 classification datasets from NanoBEIR and MTEB, respectively, and perform our analysis with the same six models used in Section~\ref{sec:k-removal}, \ref{sec:existing-theories}. 

\paragraph{Main results.}
Fig. \ref{fig:one-dim-drop} shows the results on NanoBEIR and MTEB. Each point indicates the downstream task performance achieved without the corresponding dimension. The red line is the original performance achieved by the full-sized embeddings. Points are highlighted in blue (or orange) when they are better (or worse) than the original performance. There are two main observations in the figures. (\rom{1}) In all model-benchmark combinations, a surprisingly large number of dimensions improve the performance when they are removed (blue in the figures). In other words, there are a large number of dimensions, more than half in some cases, that are degrading the embeddings' performance. In the remainder of the paper, we call them degrading dimensions. For instance, we find 498 and 398 degrading dimensions in embeddings produced by Contriever on NanoBEIR and MTEB, that is, 65\% and 52\% of the whole embeddings. (\rom{2}) The degrading dimensions are uniformly distributed across embedding dimensions without clustering in some areas.

\begin{figure}[h]
    \centering
    \small
    \includegraphics[width=1.0\columnwidth]{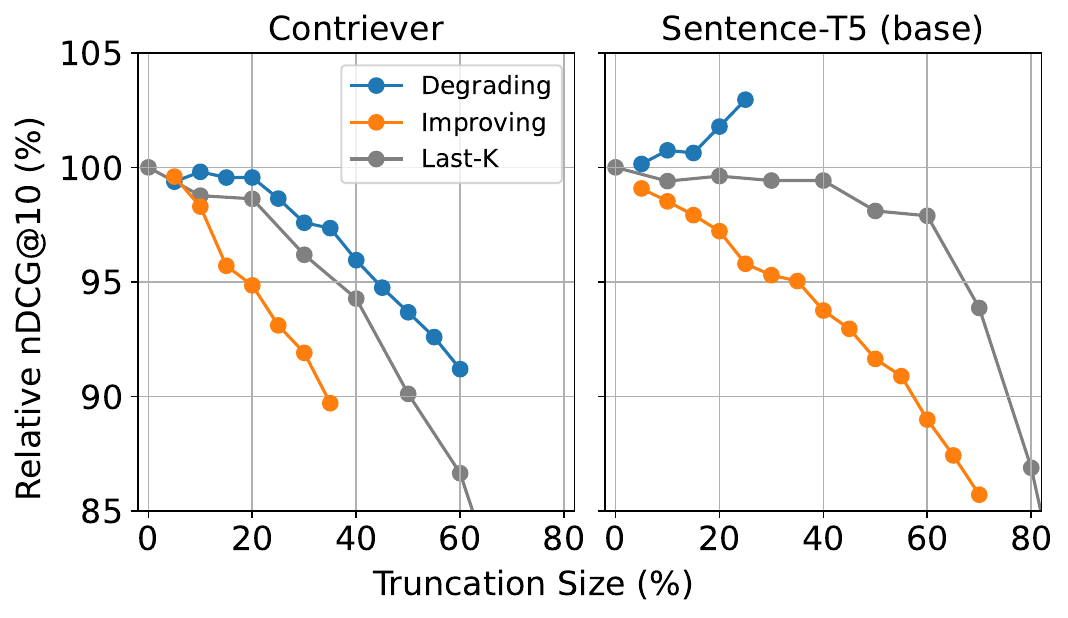}
    \caption{
    As we remove the degrading dimensions (blue plot), the relative performance for Sentence-T5 (figure on right) improves over the original embeddings. For Contriever (figure on left), while we do not see the improvements, however, the decay is slower than the last-k truncation. On the other hand, when only the improving dimensions are removed (orange plot), the performance decreases rapidly for both models. Results for other models are shown in Fig. \ref{fig:truncate-dd-id-only-others}.
    }
    \label{fig:truncate-dd-id-only-cont-and-st5}
\end{figure}

These observations provide an explanation for why removing a large number of dimensions has minimal impact on performance.
Since there are many degrading dimensions spread across embedding dimensions, random dimension removal leads to removing dimensions that both improve and degrade performance, resulting in a marginal performance drop as a whole.
We further conduct an experiment where we only remove the degrading dimensions. As the results shown in Fig. \ref{fig:truncate-dd-id-only-cont-and-st5} (blue plot), we observe that the relative performance keeps improving for some models (e.g., Sentence-T5), and even for the models whose truncated embeddings drop their performance (e.g., Contriever), the speed of decay is slower than random truncation.
Conversely, when we remove only the dimensions that are improving the performance (dimensions in orange in Fig. \ref{fig:one-dim-drop-beir}), the performance drops rapidly compared to the random truncation (Fig. \ref{fig:truncate-dd-id-only-cont-and-st5}: orange plot).

These results indicate that there is a new aspect of improvements in current text embedding models, as many of the dimensions are damaging performance. Future work can explore training objectives or model architectures that can reduce degrading dimensions to boost the model performance, similarly to how the training objective proposed by \citet{jing_understanding_2021} mitigates dimensional collapse or the novel Transformer architecture introduced by \citet{he_understanding_2024} that can reduce the number of outlier dimensions.

\paragraph{Outlier degrading dimensions in E5-Mistral.}
Some degrading dimensions in E5-Mistral's embeddings have higher impacts compared to other models when evaluated on NanoBEIR (Fig. \ref{fig:one-dim-drop-beir}). As we observe the outlier dimensions in E5-Mistral also have stronger impacts than other models (\S \ref{sec:outlier-dimensions}), we draw a connection from the outlier dimensions to the degrading dimensions.
As shown in Fig. \ref{fig:outlier-highlighted-dd}, there are 19 degrading dimensions that have stronger negative impacts than the other degrading dimensions (ODD: Outlier degrading dimensions), i.e., performance degraded at least $3\sigma$ from the mean of the performance, and 12 of them also take outlying values (ODD $\cap$ OD in the figure).
This explains our earlier observation in \S \ref{sec:outlier-dimensions}: not all outlier dimensions (OD) have a strong impact on downstream task performance, however, some indeed have outlying impacts.
We speculate that the model's extremely high relative performance (e.g., 90\% relative performance when 90\% of the dimensions are truncated) may be due to these outlier degrading dimensions.

\begin{figure}[t]
    \centering
    \small
    \includegraphics[width=1\linewidth]{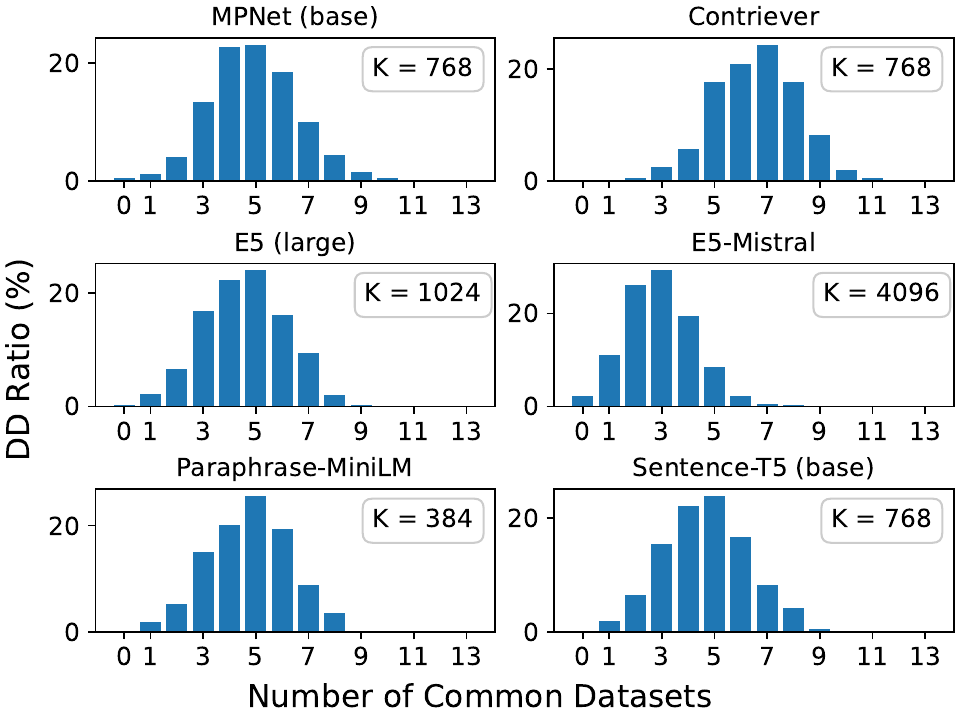}
    \caption{The ratio of degrading dimensions to the size of original embedding sizes ($K$ in the figure) that are shared across a certain number of datasets.}
    \label{fig:dd-caterorization}
\end{figure}
\paragraph{Shared degrading dimensions.}
We test if the same dimension appears as a degrading one in multiple datasets.
To this end, we independently identify degrading dimensions for each dataset in NanoBEIR and count the ones shared in multiple datasets. 
Fig. \ref{fig:dd-caterorization} shows the result.
We can see that while there is no degrading dimension shared across all 13 datasets in any models, they are indeed commonly degraded dimensions across some datasets, i.e., MPNet's more than 20\% of dimensions degrade performance in five datasets.
This observation hints at a possibility to explore domain-specific methods to identify degrading dimensions, which is left for future work.

\begin{table}[t]
    \centering
    \small
    \setlength{\tabcolsep}{3.5pt}
    \begin{tabular}{lrrrrr}
    \toprule
     & \multicolumn{2}{c}{\textbf{BEIR}} & & \multicolumn{2}{c}{\textbf{MTEB}} \\
    \cmidrule{2-3} \cmidrule{5-6}
     & \textbf{Trun. (\%)} & \textbf{PCA (\%)} & & \textbf{Trun. (\%)} & \textbf{PCA (\%)} \\
    \midrule
    \textbf{MP} & 97.7 & 99.4 & & 96.6 & 99.6 \\
    \textbf{Cont} & 87.3 & 77.3 & & 95.4 & 98.5 \\
    \textbf{E5-L} & 95.9 & 90.6 & & 97.7 & 99.8 \\
    \textbf{E5-M$^*$} & 99.6 & 100.6 & & 98.2 & 100.4 \\
    \textbf{Para} & 94.6 & 99.3 & & 97.4 & 99.5 \\
    \textbf{ST5} & 95.9 & 100.2 & & 97.8 & 99.9 \\
    \bottomrule
    \end{tabular}
    \caption{Comparison of relative performance between random truncation (Trun.) and PCA when reducing embedding size by 50\%. We use NanoBEIR for E5-Mistral.}
    \label{tab:pca_beir_mteb}
\end{table}
\paragraph{Random Truncation vs PCA.}
\label{sec:pca}
We explore the practical value of embedding truncation by comparing it to a popular dimension reduction method: PCA~\citep{raunak_effective_2019,zhang_evaluating_2024}. To this end, we compare the performance of embeddings that are halved by the PCA and the average of ten different runs of random truncation (see \S \ref{sec:k-removal}). We use the implementation from scikit-learn~\citep{pedregosa_scikitlearn_2011}, and use 20k sentences from the all-nli dataset for training\footnote{\href{https://huggingface.co/datasets/sentence-transformers/all-nli}{sentence-transformers/all-nli}}.
Table \ref{tab:pca_beir_mteb} shows the results on BEIR (NanoBEIR for E5-Mistral) and MTEB.
While random truncation does not require any training or additional computation at inference time, their relative performances are surprisingly close to PCA, even outperforming it in some cases, e.g., Contriever on BEIR.
This result exhibits random truncation as an extremely simple and cheap approach to reduce text embedding's dimensionality.

\paragraph{Truncated Representations in Causal Language Models.}
While the focus of this paper is on embeddings produced by text encoders, in this section, we study the impact that embedding truncation has on causal language modeling.

To this end, we consider two LLMs, Llama 3.1 8B \citep{grattafiori_llama_2024} and Qwen 2.5 7B \citep{qwen_qwen25_2025}, and evaluate on various six tasks after removing half of the last hidden representations before they are projected to the vocabulary space \citep{hendrycks_measuring_2020,rajpurkar_know_2018,dua_drop_2019,gordon_semeval2012_2012,cobbe_training_2021,zellers_hellaswag_2019}. We test removing the first and last half of the representations, and reduce the unembedding matrix correspondingly. We use Language Model Evaluation Harness \citep{eval-harness} for our evaluation.

\begin{table}[t]
    \centering
    \setlength{\tabcolsep}{4pt}
    \small
    \begin{tabular}{cccr}
    \toprule
    \textbf{Model} & \multicolumn{1}{c}{\textbf{Dataset}} & \multicolumn{1}{c}{\textbf{Method}} & \multicolumn{1}{c}{\textbf{Perf (Relative)}} \\
    \midrule
    \multirow{9}{*}{Llama} & \multirow{3}{*}{MMLU (acc)} & Full & 0.681 (1.000) \\
    \cmidrule{3-4}
     & & First & 0.580 (\textbf{0.852}) \\
     & & Last & 0.586 (\textbf{0.861}) \\
    \cmidrule{2-4}
     & \multirow{3}{*}{\begin{tabular}{@{}c@{}}SQUAD-V2\\(best exact)\end{tabular}} & Full & 51.87 (1.000) \\
    \cmidrule{3-4}
     & & First & 50.07 (\textbf{0.965}) \\
     & & Last & 50.07 (\textbf{0.965}) \\
    \cmidrule{2-4}
     & \multirow{3}{*}{\begin{tabular}{@{}c@{}}GSM8K\\(exact match (strict))\end{tabular}} & Full & 0.764 (1.000) \\
    \cmidrule{3-4}
     & & First & 0.009 (0.012) \\
     & & Last & 0.014 (0.018) \\
    \midrule
    \multirow{9}{*}{Qwen} & \multirow{3}{*}{MMLU (acc)} & Full & 0.718 (1.000) \\
    \cmidrule{3-4}
     & & First & 0.709 (\textbf{0.988}) \\
     & & Last & 0.709 (\textbf{0.987}) \\
    \cmidrule{2-4}
     & \multirow{3}{*}{\begin{tabular}{@{}c@{}}SQUAD-V2\\(best exact)\end{tabular}} & Full & 50.12 (1.000) \\
    \cmidrule{3-4}
     & & First & 50.07 (\textbf{0.999}) \\
     & & Last & 50.07 (\textbf{0.999}) \\
    \cmidrule{2-4}
     & \multirow{3}{*}{\begin{tabular}{@{}c@{}}GSM8K\\(exact match (strict))\end{tabular}} & Full & 0.766 (1.000)  \\
    \cmidrule{3-4}
     & & First & 0.045 (0.059) \\
     & & Last & 0.011 (0.014)\\
    \bottomrule
    \end{tabular}
    \caption{Performance on three benchmark datasets when the last hidden representations and the unembedding layer are reduced by half. Relative performance is bolded when it reaches 80\% of the original performance. The results on the rest of the datasets are shown in Table \ref{tab:llm-gen-more-datasets} in the Appendix.}
    \label{tab:llm-gen}
\end{table}
Table \ref{tab:llm-gen}, \ref{tab:llm-gen-more-datasets} shows the results. On three out of six tasks, both models retain more than 80\% of the original performance, and in these cases, similarly to our embedding-based experiments, how to reduce representations (removing first or last) does not have an impact, indicating the presence of inefficient representation space usage by LLMs. However, contrary to our embedding-based evaluation results, the high relative performance is not observed in all datasets, e.g., on GSM8K, the original performance is heavily lost with all models and reducing methods, leaving a dedicated study on LLMs for our future studies.
\section{Related Works}
\label{sec:related-works}
Inefficient use of representation space within the weights of PLMs and LLMs is hinted at by a number of papers showing these models are robust to pruning \citep{michel_are_2019,budhraja_weak_2020,chen_earlybert_2021,zheng_robust_2022}. In the context of text embeddings, several studies analyze their geometric properties, such as anisotropy \citep{hammerl_exploring_2023,godey_anisotropy_2024,razzhigaev_shape_2024}. However, their influence on downstream tasks is under exploration. \citet{ait-saada_anisotropy_2023} show a limited influence of anisotropy on text clustering. Our work adds more evidence of such limited influence of anisotropy on 26 embedding-based tasks.

The work closest to ours is \citet{kovaleva_bert_2021} in which the authors identify a few dimensions within BERT's weights that are more impactful than the others. Its successor works analyze outlier dimension in multilingual models \citep{hammerl_exploring_2023}, analyze their properties \citep{rudman_outlier_2023}, or identify similar dimensions in LLMs \citep{he_understanding_2024}. Differently, in this paper, we focus on properties of embedding dimensions instead of model weights, and while we observe outlier dimensions in embeddings, we show that they do not have a strong influence on task performance compared to the ones in model weights.

A concurrent work by \citet{tsukagoshi-sasano-2025-redundancy} shows how little dimension truncations impact performance with prompt-based text encoders.
They approach this observation from a perspective of redundancy in embeddings.
They take two concepts, namely isotropy and intrinsic dimensionality, and show that the models with higher redundancy in produced embeddings are more robust to truncation.
In addition, they show that prompt design has an influence on this phenomenon.
Our work complements this work by (\rom{1}) conducting experiments with a more diverse set of models, including non-prompt-based encoders and LLM text generators, (\rom{2}) conducting additional controlled experiments, such as the use of a continuous set of contrastively trained models to analyze redundancy (anisotropy and dimensional collapse) which allows us to have more comparable models, and finally (\rom{3}) shedding light on a new perspective to analyze the model's use of embeddings space, namely dimensional attribution analysis.
\section{Conclusion}
In this paper, we explored a surprisingly small effect of randomly removing dimensions from text embeddings on downstream task performance.
We showed that 6 text encoders can retain 90\% of the original performance even when 50\% of the dimensions are removed, consistently for 26 embedding-based downstream tasks.
Through a series of analyses, we identified a significant number of dimensions in text embeddings that are lowering downstream performance, distributed across embeddings, which would explain our initial observation.
We also observed a similar effect during the text generation by causal language models in some cases.

\section*{Limitations}
This work has the following limitations:
(\rom{1}) The cause of the degrading dimensions is still unknown. While we find that there is a large number of dimensions in text embeddings that lower the downstream task performance, this work does not explore how they emerge. For instance, identifying in which stage of model construction (e.g., early stage of pre-training) such degrading dimensions start to appear remains as future work for us.
(\rom{2}) We focused on removing one dimension at a time in our dimension attribution analysis experiments (\S \ref{sec:dimension-by-dimension}) without taking combinations of multiple dimensions into account.
While such complex analysis may provide us with more insights about degrading dimensions, this would require an extremely high computational resource, preventing us from exploring this direction.
(\rom{3}) Our experiments only cover models trained for the English language and datasets that are in English. While we confirm our findings through extensive experiments, the language is limited to English, without covering other languages, leaving a possibility of different behaviours in non-English languages.
(\rom{4}) The relation to models trained with the Matryoshka Representation Learning (MRL) method \citep{kusupati_matryoshka_2022} is not explored. The MRL models are trained with a tailored objective function so that their output representations can be truncated without large performance loss. In our preliminary experiment, we compared two MPNet variants trained with standard and MRL objectives, and observed that the performance of truncated embeddings from both models was comparable; also, the change in relative performance with different degrees of truncation was comparable to non-MRL encoders. While larger-scale experiments to assess the impact of MRL on text encoders are interesting, such experiments require powerful computational infrastructure with large GPU memories, therefore, we are unable to pursue this direction.

\section*{Acknowledgments}
The work presented in this paper is funded by the German Research Foundation (DFG) under the VADIS (PO 1900/5-1; EC 477/7-1) project, and also supported by the state of Baden-Württemberg through bwHPC and the German Research Foundation (DFG) through grant INST 35/1597-1 FUGG.

\bibliography{custom,references}

\begin{thebibliography}{83}
\providecommand{\natexlab}[1]{#1}

\bibitem[{Ait-Saada and Nadif(2023)}]{ait-saada_anisotropy_2023}
Mira Ait-Saada and Mohamed Nadif. 2023.
\newblock \href {https://doi.org/10.18653/v1/2023.acl-short.103} {Is {Anisotropy} {Truly} {Harmful}? {A} {Case} {Study} on {Text} {Clustering}}.
\newblock In \emph{Proceedings of the 61st {Annual} {Meeting} of the {Association} for {Computational} {Linguistics} ({Volume} 2: {Short} {Papers})}, pages 1194--1203, Toronto, Canada. Association for Computational Linguistics.

\bibitem[{Bastings et~al.(2022)Bastings, Ebert, Zablotskaia, Sandholm, and Filippova}]{bastings_will_2022}
Jasmijn Bastings, Sebastian Ebert, Polina Zablotskaia, Anders Sandholm, and Katja Filippova. 2022.
\newblock \href {https://doi.org/10.18653/v1/2022.emnlp-main.64} {“{Will} {You} {Find} {These} {Shortcuts}?” {A} {Protocol} for {Evaluating} the {Faithfulness} of {Input} {Salience} {Methods} for {Text} {Classification}}.
\newblock In \emph{Proceedings of the 2022 {Conference} on {Empirical} {Methods} in {Natural} {Language} {Processing}}, pages 976--991, Abu Dhabi, United Arab Emirates. Association for Computational Linguistics.

\bibitem[{Bian et~al.(2021)Bian, Huang, Cai, Yuan, and Church}]{bian_attention_2021}
Yuchen Bian, Jiaji Huang, Xingyu Cai, Jiahong Yuan, and Kenneth Church. 2021.
\newblock \href {https://aclanthology.org/2021.naacl-main.72/} {On {Attention} {Redundancy}: {A} {Comprehensive} {Study}}.
\newblock pages 930--945.

\bibitem[{Bondarenko et~al.(2020)Bondarenko, Fr{\"o}be, Beloucif, Gienapp, Ajjour, Panchenko, Biemann, Stein, Wachsmuth, Potthast et~al.}]{bondarenko2020overview}
Alexander Bondarenko, Maik Fr{\"o}be, Meriem Beloucif, Lukas Gienapp, Yamen Ajjour, Alexander Panchenko, Chris Biemann, Benno Stein, Henning Wachsmuth, Martin Potthast, and 1 others. 2020.
\newblock Overview of touch{\'e} 2020: argument retrieval.
\newblock In \emph{Experimental IR Meets Multilinguality, Multimodality, and Interaction: 11th International Conference of the CLEF Association, CLEF 2020, Thessaloniki, Greece, September 22--25, 2020, Proceedings 11}, pages 384--395. Springer.

\bibitem[{Boteva et~al.(2016)Boteva, Gholipour, Sokolov, and Riezler}]{boteva2016full}
Vera Boteva, Demian Gholipour, Artem Sokolov, and Stefan Riezler. 2016.
\newblock A full-text learning to rank dataset for medical information retrieval.
\newblock In \emph{Advances in Information Retrieval: 38th European Conference on IR Research, ECIR 2016, Padua, Italy, March 20--23, 2016. Proceedings 38}, pages 716--722. Springer.

\bibitem[{Bowman et~al.(2015)Bowman, Angeli, Potts, and Manning}]{bowman_large_2015}
Samuel~R. Bowman, Gabor Angeli, Christopher Potts, and Christopher~D. Manning. 2015.
\newblock \href {https://doi.org/10.18653/v1/D15-1075} {A large annotated corpus for learning natural language inference}.
\newblock In \emph{Proceedings of the 2015 {Conference} on {Empirical} {Methods} in {Natural} {Language} {Processing}}, pages 632--642, Lisbon, Portugal. Association for Computational Linguistics.

\bibitem[{Budhraja et~al.(2020)Budhraja, Pande, Nema, Kumar, and Khapra}]{budhraja_weak_2020}
Aakriti Budhraja, Madhura Pande, Preksha Nema, Pratyush Kumar, and Mitesh~M. Khapra. 2020.
\newblock \href {https://doi.org/10.18653/v1/2020.emnlp-main.260} {On the weak link between importance and prunability of attention heads}.
\newblock In \emph{Proceedings of the 2020 {Conference} on {Empirical} {Methods} in {Natural} {Language} {Processing} ({EMNLP})}, pages 3230--3235, Online. Association for Computational Linguistics.

\bibitem[{Casanueva et~al.(2020)Casanueva, Temčinas, Gerz, Henderson, and Vulić}]{casanueva_efficient_2020}
Iñigo Casanueva, Tadas Temčinas, Daniela Gerz, Matthew Henderson, and Ivan Vulić. 2020.
\newblock \href {https://doi.org/10.18653/v1/2020.nlp4convai-1.5} {Efficient {Intent} {Detection} with {Dual} {Sentence} {Encoders}}.
\newblock In \emph{Proceedings of the 2nd {Workshop} on {Natural} {Language} {Processing} for {Conversational} {AI}}, pages 38--45, Online. Association for Computational Linguistics.

\bibitem[{Chen et~al.(2023)Chen, Wang, Zhang, Zheng, Deng, Yu, Liu, Ma, and Zhang}]{chen_ditto_2023}
Qian Chen, Wen Wang, Qinglin Zhang, Siqi Zheng, Chong Deng, Hai Yu, Jiaqing Liu, Yukun Ma, and Chong Zhang. 2023.
\newblock \href {https://doi.org/10.18653/v1/2023.emnlp-main.359} {Ditto: {A} {Simple} and {Efficient} {Approach} to {Improve} {Sentence} {Embeddings}}.
\newblock In \emph{Proceedings of the 2023 {Conference} on {Empirical} {Methods} in {Natural} {Language} {Processing}}, pages 5868--5875, Singapore. Association for Computational Linguistics.

\bibitem[{Chen et~al.(2021)Chen, Cheng, Wang, Gan, Wang, and Liu}]{chen_earlybert_2021}
Xiaohan Chen, Yu~Cheng, Shuohang Wang, Zhe Gan, Zhangyang Wang, and Jingjing Liu. 2021.
\newblock \href {https://doi.org/10.18653/v1/2021.acl-long.171} {{EarlyBERT}: {Efficient} {BERT} training via {Early}-bird lottery tickets}.
\newblock Stroudsburg, PA, USA. Association for Computational Linguistics.

\bibitem[{cjadams et~al.(2019)cjadams, Borkan, inversion, Sorensen, Dixon, Vasserman, and nithum}]{jigsaw-unintended-bias-in-toxicity-classification}
cjadams, Daniel Borkan, inversion, Jeffrey Sorensen, Lucas Dixon, Lucy Vasserman, and nithum. 2019.
\newblock Jigsaw unintended bias in toxicity classification.
\newblock \href{https://kaggle.com/competitions/jigsaw-unintended-bias-in-toxicity-classification}{https://kaggle.com/competitions/jigsaw-unintended[...]}.
\newblock Kaggle.

\bibitem[{Cobbe et~al.(2021)Cobbe, Kosaraju, Bavarian, Chen, Jun, Kaiser, Plappert, Tworek, Hilton, Nakano, Hesse, and Schulman}]{cobbe_training_2021}
Karl Cobbe, Vineet Kosaraju, Mohammad Bavarian, Mark Chen, Heewoo Jun, Lukasz Kaiser, Matthias Plappert, Jerry Tworek, Jacob Hilton, Reiichiro Nakano, Christopher Hesse, and John Schulman. 2021.
\newblock \href {https://doi.org/10.48550/arXiv.2110.14168} {Training {Verifiers} to {Solve} {Math} {Word} {Problems}}.
\newblock \emph{arXiv preprint}.
\newblock ArXiv:2110.14168 [cs].

\bibitem[{Cohan et~al.(2020)Cohan, Feldman, Beltagy, Downey, and Weld}]{cohan_specter_2020}
Arman Cohan, Sergey Feldman, Iz~Beltagy, Doug Downey, and Daniel Weld. 2020.
\newblock \href {https://doi.org/10.18653/v1/2020.acl-main.207} {{SPECTER}: {Document}-level {Representation} {Learning} using {Citation}-informed {Transformers}}.
\newblock Stroudsburg, PA, USA. Association for Computational Linguistics.

\bibitem[{Devlin et~al.(2019)Devlin, Chang, Lee, and Toutanova}]{devlin_bert_2019}
Jacob Devlin, Ming-Wei Chang, Kenton Lee, and Kristina Toutanova. 2019.
\newblock \href {https://doi.org/10.18653/v1/N19-1423} {{BERT}: {Pre}-training of {Deep} {Bidirectional} {Transformers} for {Language} {Understanding}}.
\newblock In \emph{Proceedings of the 2019 {Conference} of the {North} {American} {Chapter} of the {Association} for {Computational} {Linguistics}: {Human} {Language} {Technologies}, {Volume} 1 ({Long} and {Short} {Papers})}, pages 4171--4186, Minneapolis, Minnesota. Association for Computational Linguistics.

\bibitem[{Dua et~al.(2019)Dua, Wang, Dasigi, Stanovsky, Singh, and Gardner}]{dua_drop_2019}
Dheeru Dua, Yizhong Wang, Pradeep Dasigi, Gabriel Stanovsky, Sameer Singh, and Matt Gardner. 2019.
\newblock \href {https://doi.org/10.18653/v1/N19-1246} {{DROP}: {A} {Reading} {Comprehension} {Benchmark} {Requiring} {Discrete} {Reasoning} {Over} {Paragraphs}}.
\newblock In \emph{Proceedings of the 2019 {Conference} of the {North} {American} {Chapter} of the {Association} for {Computational} {Linguistics}: {Human} {Language} {Technologies}, {Volume} 1 ({Long} and {Short} {Papers})}, pages 2368--2378, Minneapolis, Minnesota. Association for Computational Linguistics.

\bibitem[{FitzGerald et~al.(2023)FitzGerald, Hench, Peris, Mackie, Rottmann, Sanchez, Nash, Urbach, Kakarala, Singh, Ranganath, Crist, Britan, Leeuwis, Tur, and Natarajan}]{fitzgerald_massive_2023}
Jack FitzGerald, Christopher Hench, Charith Peris, Scott Mackie, Kay Rottmann, Ana Sanchez, Aaron Nash, Liam Urbach, Vishesh Kakarala, Richa Singh, Swetha Ranganath, Laurie Crist, Misha Britan, Wouter Leeuwis, Gokhan Tur, and Prem Natarajan. 2023.
\newblock \href {https://doi.org/10.18653/v1/2023.acl-long.235} {{MASSIVE}: {A} {1M}-{Example} {Multilingual} {Natural} {Language} {Understanding} {Dataset} with 51 {Typologically}-{Diverse} {Languages}}.
\newblock In \emph{Proceedings of the 61st {Annual} {Meeting} of the {Association} for {Computational} {Linguistics} ({Volume} 1: {Long} {Papers})}, pages 4277--4302, Toronto, Canada. Association for Computational Linguistics.

\bibitem[{Gao et~al.(2024)Gao, Tow, Abbasi, Biderman, Black, DiPofi, Foster, Golding, Hsu, Le~Noac'h, Li, McDonell, Muennighoff, Ociepa, Phang, Reynolds, Schoelkopf, Skowron, Sutawika, Tang, Thite, Wang, Wang, and Zou}]{eval-harness}
Leo Gao, Jonathan Tow, Baber Abbasi, Stella Biderman, Sid Black, Anthony DiPofi, Charles Foster, Laurence Golding, Jeffrey Hsu, Alain Le~Noac'h, Haonan Li, Kyle McDonell, Niklas Muennighoff, Chris Ociepa, Jason Phang, Laria Reynolds, Hailey Schoelkopf, Aviya Skowron, Lintang Sutawika, and 5 others. 2024.
\newblock \href {https://doi.org/10.5281/zenodo.12608602} {The language model evaluation harness}.

\bibitem[{Gao et~al.(2021)Gao, Yao, and Chen}]{gao_simcse_2021}
Tianyu Gao, Xingcheng Yao, and Danqi Chen. 2021.
\newblock \href {https://doi.org/10.18653/v1/2021.emnlp-main.552} {{SimCSE}: {Simple} {Contrastive} {Learning} of {Sentence} {Embeddings}}.
\newblock In \emph{Proceedings of the 2021 {Conference} on {Empirical} {Methods} in {Natural} {Language} {Processing}}, pages 6894--6910, Online and Punta Cana, Dominican Republic. Association for Computational Linguistics.

\bibitem[{Godey et~al.(2024)Godey, Clergerie, and Sagot}]{godey_anisotropy_2024}
Nathan Godey, Éric Clergerie, and Benoît Sagot. 2024.
\newblock \href {https://aclanthology.org/2024.eacl-long.3/} {Anisotropy {Is} {Inherent} to {Self}-{Attention} in {Transformers}}.
\newblock In \emph{Proceedings of the 18th {Conference} of the {European} {Chapter} of the {Association} for {Computational} {Linguistics} ({Volume} 1: {Long} {Papers})}, pages 35--48, St. Julian's, Malta. Association for Computational Linguistics.

\bibitem[{Gordon et~al.(2012)Gordon, Kozareva, and Roemmele}]{gordon_semeval2012_2012}
Andrew Gordon, Zornitsa Kozareva, and Melissa Roemmele. 2012.
\newblock \href {https://aclanthology.org/S12-1052/} {{SemEval}-2012 {Task} 7: {Choice} of {Plausible} {Alternatives}: {An} {Evaluation} of {Commonsense} {Causal} {Reasoning}}.
\newblock In \emph{*{SEM} 2012: {The} {First} {Joint} {Conference} on {Lexical} and {Computational} {Semantics} – {Volume} 1: {Proceedings} of the main conference and the shared task, and {Volume} 2: {Proceedings} of the {Sixth} {International} {Workshop} on {Semantic} {Evaluation} ({SemEval} 2012)}, pages 394--398, Montréal, Canada. Association for Computational Linguistics.

\bibitem[{Grattafiori et~al.(2024)Grattafiori, Dubey, Jauhri, Pandey, Kadian, Al-Dahle, Letman, Mathur, Schelten, Vaughan, Yang, Fan, Goyal, Hartshorn, Yang, Mitra, Sravankumar, Korenev, Hinsvark, Rao, Zhang, Rodriguez, Gregerson, Spataru, Roziere, Biron, Tang, Chern, Caucheteux, Nayak, Bi, Marra, McConnell, Keller, Touret, Wu, Wong, Ferrer, Nikolaidis, Allonsius, Song, Pintz, Livshits, Wyatt, Esiobu, Choudhary, Mahajan, Garcia-Olano, Perino, Hupkes, Lakomkin, AlBadawy, Lobanova, Dinan, Smith, Radenovic, Guzmán, Zhang, Synnaeve, Lee, Anderson, Thattai, Nail, Mialon, Pang, Cucurell, Nguyen, Korevaar, Xu, Touvron, Zarov, Ibarra, Kloumann, Misra, Evtimov, Zhang, Copet, Lee, Geffert, Vranes, Park, Mahadeokar, Shah, Linde, Billock, Hong, Lee, Fu, Chi, Huang, Liu, Wang, Yu, Bitton, Spisak, Park, Rocca, Johnstun, Saxe, Jia, Alwala, Prasad, Upasani, Plawiak, Li, Heafield, Stone, El-Arini, Iyer, Malik, Chiu, Bhalla, Lakhotia, Rantala-Yeary, Maaten, Chen, Tan, Jenkins, Martin, Madaan, Malo, Blecher, Landzaat,
  Oliveira, Muzzi, Pasupuleti, Singh, Paluri, Kardas, Tsimpoukelli, Oldham, Rita, Pavlova, Kambadur, Lewis, Si, Singh, Hassan, Goyal, Torabi, Bashlykov, Bogoychev, Chatterji, Zhang, Duchenne, Çelebi, Alrassy, Zhang, Li, Vasic, Weng, Bhargava, Dubal, Krishnan, Koura, Xu, He, Dong, Srinivasan, Ganapathy, Calderer, Cabral, Stojnic, Raileanu, Maheswari, Girdhar, Patel, Sauvestre, Polidoro, Sumbaly, Taylor, Silva, Hou, Wang, Hosseini, Chennabasappa, Singh, Bell, Kim, Edunov, Nie, Narang, Raparthy, Shen, Wan, Bhosale, Zhang, Vandenhende, Batra, Whitman, Sootla, Collot, Gururangan, Borodinsky, Herman, Fowler, Sheasha, Georgiou, Scialom, Speckbacher, Mihaylov, Xiao, Karn, Goswami, Gupta, Ramanathan, Kerkez, Gonguet, Do, Vogeti, Albiero, Petrovic, Chu, Xiong, Fu, Meers, Martinet, Wang, Wang, Tan, Xia, Xie, Jia, Wang, Goldschlag, Gaur, Babaei, Wen, Song, Zhang, Li, Mao, Coudert, Yan, Chen, Papakipos, Singh, Srivastava, Jain, Kelsey, Shajnfeld, Gangidi, Victoria, Goldstand, Menon, Sharma, Boesenberg, Baevski,
  Feinstein, Kallet, Sangani, Teo, Yunus, Lupu, Alvarado, Caples, Gu, Ho, Poulton, Ryan, Ramchandani, Dong, Franco, Goyal, Saraf, Chowdhury, Gabriel, Bharambe, Eisenman, Yazdan, James, Maurer, Leonhardi, Huang, Loyd, Paola, Paranjape, Liu, Wu, Ni, Hancock, Wasti, Spence, Stojkovic, Gamido, Montalvo, Parker, Burton, Mejia, Liu, Wang, Kim, Zhou, Hu, Chu, Cai, Tindal, Feichtenhofer, Gao, Civin, Beaty, Kreymer, Li, Adkins, Xu, Testuggine, David, Parikh, Liskovich, Foss, Wang, Le, Holland, Dowling, Jamil, Montgomery, Presani, Hahn, Wood, Le, Brinkman, Arcaute, Dunbar, Smothers, Sun, Kreuk, Tian, Kokkinos, Ozgenel, Caggioni, Kanayet, Seide, Florez, Schwarz, Badeer, Swee, Halpern, Herman, Sizov, Guangyi, Zhang, Lakshminarayanan, Inan, Shojanazeri, Zou, Wang, Zha, Habeeb, Rudolph, Suk, Aspegren, Goldman, Zhan, Damlaj, Molybog, Tufanov, Leontiadis, Veliche, Gat, Weissman, Geboski, Kohli, Lam, Asher, Gaya, Marcus, Tang, Chan, Zhen, Reizenstein, Teboul, Zhong, Jin, Yang, Cummings, Carvill, Shepard, McPhie, Torres,
  Ginsburg, Wang, Wu, U, Saxena, Khandelwal, Zand, Matosich, Veeraraghavan, Michelena, Li, Jagadeesh, Huang, Chawla, Huang, Chen, Garg, A, Silva, Bell, Zhang, Guo, Yu, Moshkovich, Wehrstedt, Khabsa, Avalani, Bhatt, Mankus, Hasson, Lennie, Reso, Groshev, Naumov, Lathi, Keneally, Liu, Seltzer, Valko, Restrepo, Patel, Vyatskov, Samvelyan, Clark, Macey, Wang, Hermoso, Metanat, Rastegari, Bansal, Santhanam, Parks, White, Bawa, Singhal, Egebo, Usunier, Mehta, Laptev, Dong, Cheng, Chernoguz, Hart, Salpekar, Kalinli, Kent, Parekh, Saab, Balaji, Rittner, Bontrager, Roux, Dollar, Zvyagina, Ratanchandani, Yuvraj, Liang, Alao, Rodriguez, Ayub, Murthy, Nayani, Mitra, Parthasarathy, Li, Hogan, Battey, Wang, Howes, Rinott, Mehta, Siby, Bondu, Datta, Chugh, Hunt, Dhillon, Sidorov, Pan, Mahajan, Verma, Yamamoto, Ramaswamy, Lindsay, Lindsay, Feng, Lin, Zha, Patil, Shankar, Zhang, Zhang, Wang, Agarwal, Sajuyigbe, Chintala, Max, Chen, Kehoe, Satterfield, Govindaprasad, Gupta, Deng, Cho, Virk, Subramanian, Choudhury, Goldman,
  Remez, Glaser, Best, Koehler, Robinson, Li, Zhang, Matthews, Chou, Shaked, Vontimitta, Ajayi, Montanez, Mohan, Kumar, Mangla, Ionescu, Poenaru, Mihailescu, Ivanov, Li, Wang, Jiang, Bouaziz, Constable, Tang, Wu, Wang, Wu, Gao, Kleinman, Chen, Hu, Jia, Qi, Li, Zhang, Zhang, Adi, Nam, Yu, Wang, Zhao, Hao, Qian, Li, He, Rait, DeVito, Rosnbrick, Wen, Yang, Zhao, and Ma}]{grattafiori_llama_2024}
Aaron Grattafiori, Abhimanyu Dubey, Abhinav Jauhri, Abhinav Pandey, Abhishek Kadian, Ahmad Al-Dahle, Aiesha Letman, Akhil Mathur, Alan Schelten, Alex Vaughan, Amy Yang, Angela Fan, Anirudh Goyal, Anthony Hartshorn, Aobo Yang, Archi Mitra, Archie Sravankumar, Artem Korenev, Arthur Hinsvark, and 542 others. 2024.
\newblock \href {https://doi.org/10.48550/arXiv.2407.21783} {The {Llama} 3 {Herd} of {Models}}.
\newblock \emph{arXiv preprint}.
\newblock ArXiv:2407.21783 [cs].

\bibitem[{Hasibi et~al.(2017)Hasibi, Nikolaev, Xiong, Balog, Bratsberg, Kotov, and Callan}]{dbpedia}
Faegheh Hasibi, Fedor Nikolaev, Chenyan Xiong, Krisztian Balog, Svein~Erik Bratsberg, Alexander Kotov, and Jamie Callan. 2017.
\newblock \href {https://doi.org/10.1145/3077136.3080751} {Dbpedia-entity v2: A test collection for entity search}.
\newblock In \emph{Proceedings of the 40th International ACM SIGIR Conference on Research and Development in Information Retrieval}, SIGIR '17, page 1265–1268, New York, NY, USA. Association for Computing Machinery.

\bibitem[{He et~al.(2024)He, Noci, Paliotta, Schlag, and Hofmann}]{he_understanding_2024}
Bobby He, Lorenzo Noci, Daniele Paliotta, Imanol Schlag, and Thomas Hofmann. 2024.
\newblock \href {https://proceedings.neurips.cc/paper_files/paper/2024/hash/986292a930c3692168b177a770025ab3-Abstract-Conference.html} {Understanding and {Minimising} {Outlier} {Features} in {Transformer} {Training}}.
\newblock \emph{Advances in Neural Information Processing Systems}, 37:83786--83846.

\bibitem[{He and Ozay(2022)}]{he_exploring_2022}
Bobby He and Mete Ozay. 2022.
\newblock \href {https://proceedings.mlr.press/v162/he22c.html} {Exploring the {Gap} between {Collapsed} \& {Whitened} {Features} in {Self}-{Supervised} {Learning}}.
\newblock In \emph{Proceedings of the 39th {International} {Conference} on {Machine} {Learning}}, pages 8613--8634. PMLR.
\newblock ISSN: 2640-3498.

\bibitem[{Hendrycks et~al.(2020)Hendrycks, Burns, Basart, Zou, Mazeika, Song, and Steinhardt}]{hendrycks_measuring_2020}
Dan Hendrycks, Collin Burns, Steven Basart, Andy Zou, Mantas Mazeika, Dawn Song, and Jacob Steinhardt. 2020.
\newblock \href {https://openreview.net/forum?id=d7KBjmI3GmQ} {Measuring {Massive} {Multitask} {Language} {Understanding}}.

\bibitem[{Hewitt and Manning(2019)}]{hewitt-manning-2019-structural}
John Hewitt and Christopher~D. Manning. 2019.
\newblock \href {https://doi.org/10.18653/v1/N19-1419} {{A} structural probe for finding syntax in word representations}.
\newblock In \emph{Proceedings of the 2019 Conference of the North {A}merican Chapter of the Association for Computational Linguistics: Human Language Technologies, Volume 1 (Long and Short Papers)}, pages 4129--4138, Minneapolis, Minnesota. Association for Computational Linguistics.

\bibitem[{Hua et~al.(2021)Hua, Wang, Xue, Ren, Wang, and Zhao}]{hua_feature_2021}
Tianyu Hua, Wenxiao Wang, Zihui Xue, Sucheng Ren, Yue Wang, and Hang Zhao. 2021.
\newblock \href {https://doi.org/10.1109/iccv48922.2021.00946} {On {Feature} {Decorrelation} in {Self}-{Supervised} {Learning}}.
\newblock In \emph{2021 {IEEE}/{CVF} {International} {Conference} on {Computer} {Vision} ({ICCV})}, pages 9578--9588, Montreal, QC, Canada. IEEE.

\bibitem[{Huang et~al.(2023)Huang, Campello, Erfani, Ma, Houle, and Bailey}]{huang_ldreg_2023}
Hanxun Huang, Ricardo J. G.~B. Campello, Sarah~Monazam Erfani, Xingjun Ma, Michael~E. Houle, and James Bailey. 2023.
\newblock \href {https://openreview.net/forum?id=oZyAqjAjJW&noteId=oZyAqjAjJW} {{LDReg}: {Local} {Dimensionality} {Regularized} {Self}-{Supervised} {Learning}}.

\bibitem[{Hämmerl et~al.(2023)Hämmerl, Fastowski, Libovický, and Fraser}]{hammerl_exploring_2023}
Katharina Hämmerl, Alina Fastowski, Jindřich Libovický, and Alexander Fraser. 2023.
\newblock \href {https://doi.org/10.18653/v1/2023.findings-acl.439} {Exploring {Anisotropy} and {Outliers} in {Multilingual} {Language} {Models} for {Cross}-{Lingual} {Semantic} {Sentence} {Similarity}}.
\newblock In \emph{Findings of the {Association} for {Computational} {Linguistics}: {ACL} 2023}, pages 7023--7037, Toronto, Canada. Association for Computational Linguistics.

\bibitem[{Izacard et~al.(2022)Izacard, Caron, Hosseini, Riedel, Bojanowski, Joulin, and Grave}]{izacard_unsupervised_2022}
Gautier Izacard, Mathilde Caron, Lucas Hosseini, Sebastian Riedel, Piotr Bojanowski, Armand Joulin, and Edouard Grave. 2022.
\newblock \href {https://openreview.net/forum?id=jKN1pXi7b0} {Unsupervised {Dense} {Information} {Retrieval} with {Contrastive} {Learning}}.
\newblock \emph{Transactions on Machine Learning Research}.

\bibitem[{Jing et~al.(2021)Jing, Vincent, LeCun, and Tian}]{jing_understanding_2021}
Li~Jing, Pascal Vincent, Yann LeCun, and Yuandong Tian. 2021.
\newblock \href {https://openreview.net/forum?id=YevsQ05DEN7} {Understanding {Dimensional} {Collapse} in {Contrastive} {Self}-supervised {Learning}}.

\bibitem[{Karpukhin et~al.(2020)Karpukhin, Oguz, Min, Lewis, Wu, Edunov, Chen, and Yih}]{karpukhin-etal-2020-dense}
Vladimir Karpukhin, Barlas Oguz, Sewon Min, Patrick Lewis, Ledell Wu, Sergey Edunov, Danqi Chen, and Wen-tau Yih. 2020.
\newblock \href {https://doi.org/10.18653/v1/2020.emnlp-main.550} {Dense passage retrieval for open-domain question answering}.
\newblock In \emph{Proceedings of the 2020 Conference on Empirical Methods in Natural Language Processing (EMNLP)}, pages 6769--6781, Online. Association for Computational Linguistics.

\bibitem[{Keung et~al.(2020)Keung, Lu, Szarvas, and Smith}]{keung_multilingual_2020}
Phillip Keung, Yichao Lu, György Szarvas, and Noah~A. Smith. 2020.
\newblock \href {https://doi.org/10.18653/v1/2020.emnlp-main.369} {The {Multilingual} {Amazon} {Reviews} {Corpus}}.
\newblock In \emph{Proceedings of the 2020 {Conference} on {Empirical} {Methods} in {Natural} {Language} {Processing} ({EMNLP})}, pages 4563--4568, Online. Association for Computational Linguistics.

\bibitem[{Kovaleva et~al.(2021)Kovaleva, Kulshreshtha, Rogers, and Rumshisky}]{kovaleva_bert_2021}
Olga Kovaleva, Saurabh Kulshreshtha, Anna Rogers, and Anna Rumshisky. 2021.
\newblock \href {https://doi.org/10.18653/v1/2021.findings-acl.300} {{BERT} {Busters}: {Outlier} {Dimensions} that {Disrupt} {Transformers}}.
\newblock In \emph{Findings of the {Association} for {Computational} {Linguistics}: {ACL}-{IJCNLP} 2021}, pages 3392--3405, Online. Association for Computational Linguistics.

\bibitem[{Kulmizev et~al.(2020)Kulmizev, Ravishankar, Abdou, and Nivre}]{kulmizev-etal-2020-neural}
Artur Kulmizev, Vinit Ravishankar, Mostafa Abdou, and Joakim Nivre. 2020.
\newblock \href {https://doi.org/10.18653/v1/2020.acl-main.375} {Do neural language models show preferences for syntactic formalisms?}
\newblock In \emph{Proceedings of the 58th Annual Meeting of the Association for Computational Linguistics}, pages 4077--4091, Online. Association for Computational Linguistics.

\bibitem[{Kusupati et~al.(2022)Kusupati, Bhatt, Rege, Wallingford, Sinha, Ramanujan, Howard-Snyder, Chen, Kakade, Jain, and Farhadi}]{kusupati_matryoshka_2022}
Aditya Kusupati, Gantavya Bhatt, Aniket Rege, Matthew Wallingford, Aditya Sinha, Vivek Ramanujan, William Howard-Snyder, Kaifeng Chen, Sham Kakade, Prateek Jain, and Ali Farhadi. 2022.
\newblock \href {https://papers.nips.cc/paper_files/paper/2022/hash/c32319f4868da7613d78af9993100e42-Abstract-Conference.html} {Matryoshka {Representation} {Learning}}.
\newblock \emph{Advances in Neural Information Processing Systems}, 35:30233--30249.

\bibitem[{Kwiatkowski et~al.(2019)Kwiatkowski, Palomaki, Redfield, Collins, Parikh, Alberti, Epstein, Polosukhin, Devlin, Lee, Toutanova, Jones, Kelcey, Chang, Dai, Uszkoreit, Le, and Petrov}]{kwiatkowski_natural_2019}
Tom Kwiatkowski, Jennimaria Palomaki, Olivia Redfield, Michael Collins, Ankur Parikh, Chris Alberti, Danielle Epstein, Illia Polosukhin, Jacob Devlin, Kenton Lee, Kristina Toutanova, Llion Jones, Matthew Kelcey, Ming-Wei Chang, Andrew~M. Dai, Jakob Uszkoreit, Quoc Le, and Slav Petrov. 2019.
\newblock \href {https://doi.org/10.1162/tacl_a_00276} {Natural {Questions}: {A} {Benchmark} for {Question} {Answering} {Research}}.
\newblock \emph{Transactions of the Association for Computational Linguistics}, 7:453--466.

\bibitem[{Leippold and Diggelmann(2020)}]{leippold2020climatefever}
Markus Leippold and Thomas Diggelmann. 2020.
\newblock \href {https://www.climatechange.ai/papers/neurips2020/67} {Climate-fever: A dataset for verification of real-world climate claims}.
\newblock In \emph{NeurIPS 2020 Workshop on Tackling Climate Change with Machine Learning}.

\bibitem[{Li et~al.(2021)Li, Arora, Chen, Gupta, Gupta, and Mehdad}]{li_mtop_2021}
Haoran Li, Abhinav Arora, Shuohui Chen, Anchit Gupta, Sonal Gupta, and Yashar Mehdad. 2021.
\newblock \href {https://doi.org/10.18653/v1/2021.eacl-main.257} {{MTOP}: {A} {Comprehensive} {Multilingual} {Task}-{Oriented} {Semantic} {Parsing} {Benchmark}}.
\newblock In \emph{Proceedings of the 16th {Conference} of the {European} {Chapter} of the {Association} for {Computational} {Linguistics}: {Main} {Volume}}, pages 2950--2962, Online. Association for Computational Linguistics.

\bibitem[{Li et~al.(2025)Li, Stenzel, Eickhoff, and Bahrainian}]{li-etal-2025-enhancing-retrieval}
Siran Li, Linus Stenzel, Carsten Eickhoff, and Seyed~Ali Bahrainian. 2025.
\newblock \href {https://aclanthology.org/2025.coling-main.449/} {Enhancing retrieval-augmented generation: A study of best practices}.
\newblock In \emph{Proceedings of the 31st International Conference on Computational Linguistics}, pages 6705--6717, Abu Dhabi, UAE. Association for Computational Linguistics.

\bibitem[{Liu et~al.(2021)Liu, Wang, Wang, Ye, Xi, and Zhang}]{liu-etal-2021-improving-embedding-based}
Peiyang Liu, Xi~Wang, Sen Wang, Wei Ye, Xiangyu Xi, and Shikun Zhang. 2021.
\newblock \href {https://doi.org/10.18653/v1/2021.findings-emnlp.13} {Improving embedding-based large-scale retrieval via label enhancement}.
\newblock In \emph{Findings of the Association for Computational Linguistics: EMNLP 2021}, pages 133--142, Punta Cana, Dominican Republic. Association for Computational Linguistics.

\bibitem[{Maas et~al.(2011)Maas, Daly, Pham, Huang, Ng, and Potts}]{maas_learning_2011}
Andrew~L. Maas, Raymond~E. Daly, Peter~T. Pham, Dan Huang, Andrew~Y. Ng, and Christopher Potts. 2011.
\newblock \href {https://aclanthology.org/P11-1015/} {Learning {Word} {Vectors} for {Sentiment} {Analysis}}.
\newblock In \emph{Proceedings of the 49th {Annual} {Meeting} of the {Association} for {Computational} {Linguistics}: {Human} {Language} {Technologies}}, pages 142--150, Portland, Oregon, USA. Association for Computational Linguistics.

\bibitem[{Maggie et~al.(2020)Maggie, Culliton, and Chen}]{tweet-sentiment-extraction}
Maggie, Phil Culliton, and Wei Chen. 2020.
\newblock Tweet sentiment extraction.
\newblock \url{https://kaggle.com/competitions/tweet-sentiment-extraction}.
\newblock Kaggle.

\bibitem[{Maia et~al.(2018)Maia, Handschuh, Freitas, Davis, McDermott, Zarrouk, and Balahur}]{fiqa}
Macedo Maia, Siegfried Handschuh, Andr\'{e} Freitas, Brian Davis, Ross McDermott, Manel Zarrouk, and Alexandra Balahur. 2018.
\newblock \href {https://doi.org/10.1145/3184558.3192301} {Www'18 open challenge: Financial opinion mining and question answering}.
\newblock In \emph{Companion Proceedings of the The Web Conference 2018}, WWW '18, page 1941–1942, Republic and Canton of Geneva, CHE. International World Wide Web Conferences Steering Committee.

\bibitem[{McAuley and Leskovec(2013)}]{10.1145/2507157.2507163}
Julian McAuley and Jure Leskovec. 2013.
\newblock \href {https://doi.org/10.1145/2507157.2507163} {Hidden factors and hidden topics: understanding rating dimensions with review text}.
\newblock In \emph{Proceedings of the 7th ACM Conference on Recommender Systems}, RecSys '13, page 165–172, New York, NY, USA. Association for Computing Machinery.

\bibitem[{Michel et~al.(2019)Michel, Levy, and Neubig}]{michel_are_2019}
Paul Michel, Omer Levy, and Graham Neubig. 2019.
\newblock \href {https://papers.nips.cc/paper_files/paper/2019/hash/2c601ad9d2ff9bc8b282670cdd54f69f-Abstract.html} {Are {Sixteen} {Heads} {Really} {Better} than {One}?}
\newblock In \emph{Advances in {Neural} {Information} {Processing} {Systems}}, volume~32. Curran Associates, Inc.

\bibitem[{Muennighoff et~al.(2023)Muennighoff, Tazi, Magne, and Reimers}]{muennighoff_mteb_2023}
Niklas Muennighoff, Nouamane Tazi, Loic Magne, and Nils Reimers. 2023.
\newblock \href {https://doi.org/10.18653/v1/2023.eacl-main.148} {{MTEB}: {Massive} {Text} {Embedding} {Benchmark}}.
\newblock In \emph{Proceedings of the 17th {Conference} of the {European} {Chapter} of the {Association} for {Computational} {Linguistics}}, pages 2014--2037, Dubrovnik, Croatia. Association for Computational Linguistics.

\bibitem[{Nguyen et~al.(2016)Nguyen, Rosenberg, Song, Gao, Tiwary, Majumder, and Deng}]{nguyen2016ms}
Tri Nguyen, Mir Rosenberg, Xia Song, Jianfeng Gao, Saurabh Tiwary, Rangan Majumder, and Li~Deng. 2016.
\newblock \href {https://www.microsoft.com/en-us/research/publication/ms-marco-human-generated-machine-reading-comprehension-dataset/} {Ms marco: A human generated machine reading comprehension dataset}.

\bibitem[{Ni et~al.(2022)Ni, Hernandez~Abrego, Constant, Ma, Hall, Cer, and Yang}]{ni_sentencet5_2022}
Jianmo Ni, Gustavo Hernandez~Abrego, Noah Constant, Ji~Ma, Keith Hall, Daniel Cer, and Yinfei Yang. 2022.
\newblock \href {https://doi.org/10.18653/v1/2022.findings-acl.146} {Sentence-{T5}: {Scalable} sentence encoders from pre-trained text-to-text models}.
\newblock In \emph{Findings of the association for computational linguistics: {ACL} 2022}, pages 1864--1874, Dublin, Ireland. Association for Computational Linguistics.

\bibitem[{O'Neill et~al.(2021)O'Neill, Rozenshtein, Kiryo, Kubota, and Bollegala}]{oneill_wish_2021}
James O'Neill, Polina Rozenshtein, Ryuichi Kiryo, Motoko Kubota, and Danushka Bollegala. 2021.
\newblock \href {https://doi.org/10.18653/v1/2021.emnlp-main.568} {I {Wish} {I} {Would} {Have} {Loved} {This} {One}, {But} {I} {Didn}`t – {A} {Multilingual} {Dataset} for {Counterfactual} {Detection} in {Product} {Review}}.
\newblock In \emph{Proceedings of the 2021 {Conference} on {Empirical} {Methods} in {Natural} {Language} {Processing}}, pages 7092--7108, Online and Punta Cana, Dominican Republic. Association for Computational Linguistics.

\bibitem[{Oord et~al.(2019)Oord, Li, and Vinyals}]{oord_representation_2019}
Aaron van~den Oord, Yazhe Li, and Oriol Vinyals. 2019.
\newblock \href {https://doi.org/10.48550/arXiv.1807.03748} {Representation {Learning} with {Contrastive} {Predictive} {Coding}}.
\newblock \emph{arXiv preprint}.
\newblock ArXiv:1807.03748 [cs].

\bibitem[{Pedregosa et~al.(2011)Pedregosa, Varoquaux, Gramfort, Michel, Thirion, Grisel, Blondel, Prettenhofer, Weiss, Dubourg, Vanderplas, Passos, Cournapeau, Brucher, Perrot, and Duchesnay}]{pedregosa_scikitlearn_2011}
Fabian Pedregosa, Gaël Varoquaux, Alexandre Gramfort, Vincent Michel, Bertrand Thirion, Olivier Grisel, Mathieu Blondel, Peter Prettenhofer, Ron Weiss, Vincent Dubourg, Jake Vanderplas, Alexandre Passos, David Cournapeau, Matthieu Brucher, Matthieu Perrot, and Édouard Duchesnay. 2011.
\newblock \href {http://jmlr.org/papers/v12/pedregosa11a.html} {Scikit-learn: {Machine} {Learning} in {Python}}.
\newblock \emph{Journal of Machine Learning Research}, 12(85):2825--2830.

\bibitem[{Puccetti et~al.(2022)Puccetti, Rogers, Drozd, and Dell'Orletta}]{puccetti_outlier_2022}
Giovanni Puccetti, Anna Rogers, Aleksandr Drozd, and Felice Dell'Orletta. 2022.
\newblock \href {https://doi.org/10.18653/v1/2022.findings-emnlp.93} {Outlier {Dimensions} that {Disrupt} {Transformers} are {Driven} by {Frequency}}.
\newblock In \emph{Findings of the {Association} for {Computational} {Linguistics}: {EMNLP} 2022}, pages 1286--1304, Abu Dhabi, United Arab Emirates. Association for Computational Linguistics.

\bibitem[{Qwen et~al.(2025)Qwen, Yang, Yang, Zhang, Hui, Zheng, Yu, Li, Liu, Huang, Wei, Lin, Yang, Tu, Zhang, Yang, Yang, Zhou, Lin, Dang, Lu, Bao, Yang, Yu, Li, Xue, Zhang, Zhu, Men, Lin, Li, Tang, Xia, Ren, Ren, Fan, Su, Zhang, Wan, Liu, Cui, Zhang, and Qiu}]{qwen_qwen25_2025}
Qwen, An~Yang, Baosong Yang, Beichen Zhang, Binyuan Hui, Bo~Zheng, Bowen Yu, Chengyuan Li, Dayiheng Liu, Fei Huang, Haoran Wei, Huan Lin, Jian Yang, Jianhong Tu, Jianwei Zhang, Jianxin Yang, Jiaxi Yang, Jingren Zhou, Junyang Lin, and 24 others. 2025.
\newblock \href {https://doi.org/10.48550/arXiv.2412.15115} {Qwen2.5 {Technical} {Report}}.
\newblock \emph{arXiv preprint}.
\newblock ArXiv:2412.15115 [cs].

\bibitem[{Raffel et~al.(2020)Raffel, Shazeer, Roberts, Lee, Narang, Matena, Zhou, Li, and Liu}]{raffel_exploring_2020}
Colin Raffel, Noam Shazeer, Adam Roberts, Katherine Lee, Sharan Narang, Michael Matena, Yanqi Zhou, Wei Li, and Peter~J Liu. 2020.
\newblock \href {https://jmlr.org/papers/v21/20-074.html} {Exploring the {Limits} of {Transfer} {Learning} with a {Unified} {Text}-to-{Text} {Transformer}}.
\newblock \emph{J. Mach. Learn. Res.}, 21(140):1--67.

\bibitem[{Rajpurkar et~al.(2018)Rajpurkar, Jia, and Liang}]{rajpurkar_know_2018}
Pranav Rajpurkar, Robin Jia, and Percy Liang. 2018.
\newblock \href {https://doi.org/10.18653/v1/P18-2124} {Know {What} {You} {Don}`t {Know}: {Unanswerable} {Questions} for {SQuAD}}.
\newblock In \emph{Proceedings of the 56th {Annual} {Meeting} of the {Association} for {Computational} {Linguistics} ({Volume} 2: {Short} {Papers})}, pages 784--789, Melbourne, Australia. Association for Computational Linguistics.

\bibitem[{Raunak et~al.(2019)Raunak, Gupta, and Metze}]{raunak_effective_2019}
Vikas Raunak, Vivek Gupta, and Florian Metze. 2019.
\newblock \href {https://doi.org/10.18653/v1/W19-4328} {Effective {Dimensionality} {Reduction} for {Word} {Embeddings}}.
\newblock In \emph{Proceedings of the 4th {Workshop} on {Representation} {Learning} for {NLP} ({RepL4NLP}-2019)}, pages 235--243, Florence, Italy. Association for Computational Linguistics.

\bibitem[{Razzhigaev et~al.(2024)Razzhigaev, Mikhalchuk, Goncharova, Oseledets, Dimitrov, and Kuznetsov}]{razzhigaev_shape_2024}
Anton Razzhigaev, Matvey Mikhalchuk, Elizaveta Goncharova, Ivan Oseledets, Denis Dimitrov, and Andrey Kuznetsov. 2024.
\newblock \href {https://aclanthology.org/2024.findings-eacl.58/} {The {Shape} of {Learning}: {Anisotropy} and {Intrinsic} {Dimensions} in {Transformer}-{Based} {Models}}.
\newblock In \emph{Findings of the {Association} for {Computational} {Linguistics}: {EACL} 2024}, pages 868--874, St. Julian's, Malta. Association for Computational Linguistics.

\bibitem[{Reimers and Gurevych(2019)}]{reimers_sentencebert_2019}
Nils Reimers and Iryna Gurevych. 2019.
\newblock \href {https://doi.org/10.18653/v1/D19-1410} {Sentence-{BERT}: {Sentence} {Embeddings} using {Siamese} {BERT}-{Networks}}.
\newblock In \emph{Proceedings of the 2019 {Conference} on {Empirical} {Methods} in {Natural} {Language} {Processing} and the 9th {International} {Joint} {Conference} on {Natural} {Language} {Processing} ({EMNLP}-{IJCNLP})}, pages 3982--3992, Hong Kong, China. Association for Computational Linguistics.

\bibitem[{Rudman et~al.(2023)Rudman, Chen, and Eickhoff}]{rudman_outlier_2023}
William Rudman, Catherine Chen, and Carsten Eickhoff. 2023.
\newblock \href {https://doi.org/10.18653/v1/2023.emnlp-main.901} {Outlier {Dimensions} {Encode} {Task} {Specific} {Knowledge}}.
\newblock In \emph{Proceedings of the 2023 {Conference} on {Empirical} {Methods} in {Natural} {Language} {Processing}}, pages 14596--14605, Singapore. Association for Computational Linguistics.

\bibitem[{Rudman and Eickhoff(2023)}]{rudman_stable_2023}
William Rudman and Carsten Eickhoff. 2023.
\newblock \href {https://openreview.net/forum?id=dbQH9AOVd5} {Stable {Anisotropic} {Regularization}}.

\bibitem[{Rudman et~al.(2022)Rudman, Gillman, Rayne, and Eickhoff}]{rudman_isoscore_2022}
William Rudman, Nate Gillman, Taylor Rayne, and Carsten Eickhoff. 2022.
\newblock \href {https://doi.org/10.18653/v1/2022.findings-acl.262} {{IsoScore}: {Measuring} the {Uniformity} of {Embedding} {Space} {Utilization}}.
\newblock In \emph{Findings of the {Association} for {Computational} {Linguistics}: {ACL} 2022}, pages 3325--3339, Dublin, Ireland. Association for Computational Linguistics.

\bibitem[{Saravia et~al.(2018)Saravia, Liu, Huang, Wu, and Chen}]{saravia_carer_2018}
Elvis Saravia, Hsien-Chi~Toby Liu, Yen-Hao Huang, Junlin Wu, and Yi-Shin Chen. 2018.
\newblock \href {https://doi.org/10.18653/v1/D18-1404} {{CARER}: {Contextualized} {Affect} {Representations} for {Emotion} {Recognition}}.
\newblock In \emph{Proceedings of the 2018 {Conference} on {Empirical} {Methods} in {Natural} {Language} {Processing}}, pages 3687--3697, Brussels, Belgium. Association for Computational Linguistics.

\bibitem[{Sattarzadeh et~al.(2021)Sattarzadeh, Sudhakar, Lem, Mehryar, Plataniotis, Jang, Kim, Jeong, Lee, and Bae}]{sattarzadeh_explaining_2021}
Sam Sattarzadeh, Mahesh Sudhakar, Anthony Lem, Shervin Mehryar, Konstantinos~N. Plataniotis, Jongseong Jang, Hyunwoo Kim, Yeonjeong Jeong, Sangmin Lee, and Kyunghoon Bae. 2021.
\newblock \href {https://doi.org/10.1609/aaai.v35i13.17384} {Explaining {Convolutional} {Neural} {Networks} through {Attribution}-{Based} {Input} {Sampling} and {Block}-{Wise} {Feature} {Aggregation}}.
\newblock \emph{Proceedings of the AAAI Conference on Artificial Intelligence}, 35(13):11639--11647.
\newblock Number: 13.

\bibitem[{Serrano and Smith(2019)}]{serrano_attention_2019}
Sofia Serrano and Noah~A. Smith. 2019.
\newblock \href {https://doi.org/10.18653/v1/P19-1282} {Is {Attention} {Interpretable}?}
\newblock In \emph{Proceedings of the 57th {Annual} {Meeting} of the {Association} for {Computational} {Linguistics}}, pages 2931--2951, Florence, Italy. Association for Computational Linguistics.

\bibitem[{Sundararajan et~al.(2017)Sundararajan, Taly, and Yan}]{sundararajan_axiomatic_2017}
Mukund Sundararajan, Ankur Taly, and Qiqi Yan. 2017.
\newblock \href {https://proceedings.mlr.press/v70/sundararajan17a.html} {Axiomatic {Attribution} for {Deep} {Networks}}.
\newblock In \emph{Proceedings of the 34th {International} {Conference} on {Machine} {Learning}}, pages 3319--3328. PMLR.
\newblock ISSN: 2640-3498.

\bibitem[{Thakur et~al.(2021)Thakur, Reimers, Rücklé, Srivastava, and Gurevych}]{thakur_beir_2021}
Nandan Thakur, Nils Reimers, Andreas Rücklé, Abhishek Srivastava, and Iryna Gurevych. 2021.
\newblock \href {https://openreview.net/forum?id=wCu6T5xFjeJ} {{BEIR}: {A} {Heterogeneous} {Benchmark} for {Zero}-shot {Evaluation} of {Information} {Retrieval} {Models}}.

\bibitem[{Thorne et~al.(2018)Thorne, Vlachos, Christodoulopoulos, and Mittal}]{thorne_fever_2018}
James Thorne, Andreas Vlachos, Christos Christodoulopoulos, and Arpit Mittal. 2018.
\newblock \href {https://doi.org/10.18653/v1/N18-1074} {{FEVER}: a {Large}-scale {Dataset} for {Fact} {Extraction} and {VERification}}.
\newblock In \emph{Proceedings of the 2018 {Conference} of the {North} {American} {Chapter} of the {Association} for {Computational} {Linguistics}: {Human} {Language} {Technologies}, {Volume} 1 ({Long} {Papers})}, pages 809--819, New Orleans, Louisiana. Association for Computational Linguistics.

\bibitem[{Tsukagoshi and Sasano(2025)}]{tsukagoshi-sasano-2025-redundancy}
Hayato Tsukagoshi and Ryohei Sasano. 2025.
\newblock \href {https://doi.org/10.18653/v1/2025.findings-acl.1330} {Redundancy, isotropy, and intrinsic dimensionality of prompt-based text embeddings}.
\newblock In \emph{Findings of the Association for Computational Linguistics: ACL 2025}, pages 25915--25930, Vienna, Austria. Association for Computational Linguistics.

\bibitem[{Voorhees et~al.(2021)Voorhees, Alam, Bedrick, Demner-Fushman, Hersh, Lo, Roberts, Soboroff, and Wang}]{TREC-COVID}
Ellen Voorhees, Tasmeer Alam, Steven Bedrick, Dina Demner-Fushman, William~R. Hersh, Kyle Lo, Kirk Roberts, Ian Soboroff, and Lucy~Lu Wang. 2021.
\newblock \href {https://doi.org/10.1145/3451964.3451965} {Trec-covid: constructing a pandemic information retrieval test collection}.
\newblock \emph{SIGIR Forum}, 54(1).

\bibitem[{Wachsmuth et~al.(2018)Wachsmuth, Syed, and Stein}]{wachsmuth_retrieval_2018}
Henning Wachsmuth, Shahbaz Syed, and Benno Stein. 2018.
\newblock \href {https://doi.org/10.18653/v1/P18-1023} {Retrieval of the {Best} {Counterargument} without {Prior} {Topic} {Knowledge}}.
\newblock In \emph{Proceedings of the 56th {Annual} {Meeting} of the {Association} for {Computational} {Linguistics} ({Volume} 1: {Long} {Papers})}, pages 241--251, Melbourne, Australia. Association for Computational Linguistics.

\bibitem[{Wadden et~al.(2020)Wadden, Lin, Lo, Wang, van Zuylen, Cohan, and Hajishirzi}]{wadden_fact_2020}
David Wadden, Shanchuan Lin, Kyle Lo, Lucy~Lu Wang, Madeleine van Zuylen, Arman Cohan, and Hannaneh Hajishirzi. 2020.
\newblock \href {https://doi.org/10.18653/v1/2020.emnlp-main.609} {Fact or fiction: {Verifying} scientific claims}.
\newblock Stroudsburg, PA, USA. Association for Computational Linguistics.

\bibitem[{Wang et~al.(2022)Wang, Yang, Huang, Jiao, Yang, Jiang, Majumder, and Wei}]{wang_text_2022}
Liang Wang, Nan Yang, Xiaolong Huang, Binxing Jiao, Linjun Yang, Daxin Jiang, Rangan Majumder, and Furu Wei. 2022.
\newblock \href {http://arxiv.org/abs/2212.03533} {Text {Embeddings} by {Weakly}-{Supervised} {Contrastive} {Pre}-training}.
\newblock \emph{arXiv preprint}.
\newblock ArXiv:2212.03533 [cs].

\bibitem[{Wang et~al.(2024)Wang, Yang, Huang, Yang, Majumder, and Wei}]{wang_improving_2024}
Liang Wang, Nan Yang, Xiaolong Huang, Linjun Yang, Rangan Majumder, and Furu Wei. 2024.
\newblock \href {https://doi.org/10.18653/v1/2024.acl-long.642} {Improving {Text} {Embeddings} with {Large} {Language} {Models}}.
\newblock In \emph{Proceedings of the 62nd {Annual} {Meeting} of the {Association} for {Computational} {Linguistics} ({Volume} 1: {Long} {Papers})}, pages 11897--11916, Bangkok, Thailand. Association for Computational Linguistics.

\bibitem[{Wang and Isola(2020)}]{wang_understanding_2020}
Tongzhou Wang and Phillip Isola. 2020.
\newblock Understanding contrastive representation learning through alignment and uniformity on the hypersphere.
\newblock In \emph{Proceedings of the 37th international conference on machine learning}, {ICML}'20. JMLR.org.
\newblock Number of pages: 11 tex.articleno: 921.

\bibitem[{Williams et~al.(2018)Williams, Nangia, and Bowman}]{williams_broadcoverage_2018}
Adina Williams, Nikita Nangia, and Samuel Bowman. 2018.
\newblock \href {https://doi.org/10.18653/v1/N18-1101} {A {Broad}-{Coverage} {Challenge} {Corpus} for {Sentence} {Understanding} through {Inference}}.
\newblock In \emph{Proceedings of the 2018 {Conference} of the {North} {American} {Chapter} of the {Association} for {Computational} {Linguistics}: {Human} {Language} {Technologies}, {Volume} 1 ({Long} {Papers})}, pages 1112--1122, New Orleans, Louisiana. Association for Computational Linguistics.

\bibitem[{Wu et~al.(2023)Wu, Chen, Quan, Wang, and Wang}]{wu_adkd_2023}
Siyue Wu, Hongzhan Chen, Xiaojun Quan, Qifan Wang, and Rui Wang. 2023.
\newblock \href {https://doi.org/10.18653/v1/2023.acl-long.471} {{AD}-{KD}: {Attribution}-{Driven} {Knowledge} {Distillation} for {Language} {Model} {Compression}}.
\newblock In \emph{Proceedings of the 61st {Annual} {Meeting} of the {Association} for {Computational} {Linguistics} ({Volume} 1: {Long} {Papers})}, pages 8449--8465, Toronto, Canada. Association for Computational Linguistics.

\bibitem[{Xiao et~al.(2023)Xiao, Long, and Al~Moubayed}]{xiao-etal-2023-isotropy}
Chenghao Xiao, Yang Long, and Noura Al~Moubayed. 2023.
\newblock \href {https://doi.org/10.18653/v1/2023.findings-acl.778} {On isotropy, contextualization and learning dynamics of contrastive-based sentence representation learning}.
\newblock In \emph{Findings of the Association for Computational Linguistics: ACL 2023}, pages 12266--12283, Toronto, Canada. Association for Computational Linguistics.

\bibitem[{Yang et~al.(2018)Yang, Qi, Zhang, Bengio, Cohen, Salakhutdinov, and Manning}]{yang_hotpotqa_2018}
Zhilin Yang, Peng Qi, Saizheng Zhang, Yoshua Bengio, William Cohen, Ruslan Salakhutdinov, and Christopher~D. Manning. 2018.
\newblock \href {https://doi.org/10.18653/v1/D18-1259} {{HotpotQA}: {A} {Dataset} for {Diverse}, {Explainable} {Multi}-hop {Question} {Answering}}.
\newblock In \emph{Proceedings of the 2018 {Conference} on {Empirical} {Methods} in {Natural} {Language} {Processing}}, pages 2369--2380, Brussels, Belgium. Association for Computational Linguistics.

\bibitem[{Zellers et~al.(2019)Zellers, Holtzman, Bisk, Farhadi, and Choi}]{zellers_hellaswag_2019}
Rowan Zellers, Ari Holtzman, Yonatan Bisk, Ali Farhadi, and Yejin Choi. 2019.
\newblock \href {https://doi.org/10.18653/v1/P19-1472} {{HellaSwag}: {Can} a {Machine} {Really} {Finish} {Your} {Sentence}?}
\newblock In \emph{Proceedings of the 57th {Annual} {Meeting} of the {Association} for {Computational} {Linguistics}}, pages 4791--4800, Florence, Italy. Association for Computational Linguistics.

\bibitem[{Zhang et~al.(2024{\natexlab{a}})Zhang, Zhou, and Bollegala}]{zhang_evaluating_2024}
Gaifan Zhang, Yi~Zhou, and Danushka Bollegala. 2024{\natexlab{a}}.
\newblock \href {https://aclanthology.org/2024.lrec-main.579/} {Evaluating {Unsupervised} {Dimensionality} {Reduction} {Methods} for {Pretrained} {Sentence} {Embeddings}}.
\newblock In \emph{Proceedings of the 2024 {Joint} {International} {Conference} on {Computational} {Linguistics}, {Language} {Resources} and {Evaluation} ({LREC}-{COLING} 2024)}, pages 6530--6543, Torino, Italia. ELRA and ICCL.

\bibitem[{Zhang et~al.(2024{\natexlab{b}})Zhang, Yu, Zang, Eickhoff, and Pavlick}]{zhang_same_2024}
Ruochen Zhang, Qinan Yu, Matianyu Zang, Carsten Eickhoff, and Ellie Pavlick. 2024{\natexlab{b}}.
\newblock \href {https://openreview.net/forum?id=NCrFA7dq8T} {The {Same} but {Different}: {Structural} {Similarities} and {Differences} in {Multilingual} {Language} {Modeling}}.

\bibitem[{Zheng et~al.(2022)Zheng, Rong, Zhou, Liang, Wang, Wu, Gui, Zhang, and Huang}]{zheng_robust_2022}
Rui Zheng, Bao Rong, Yuhao Zhou, Di~Liang, Sirui Wang, Wei Wu, Tao Gui, Qi~Zhang, and Xuanjing Huang. 2022.
\newblock \href {https://aclanthology.org/2022.acl-long.157} {Robust {Lottery} {Tickets} for {Pre}-trained {Language} {Models}}.
\newblock In \emph{Proceedings of the 60th {Annual} {Meeting} of the {Association} for {Computational} {Linguistics} ({Volume} 1: {Long} {Papers})}, pages 2211--2224, Dublin, Ireland. Association for Computational Linguistics.

\end{thebibliography}

\appendix
\label{sec:appendix}
\newpage
\onecolumn
\section{Appendix}

\subsection{Relative Performances by Different Seeds}
\begin{figure}[h]
    \centering
    \includegraphics[width=0.8\textwidth]{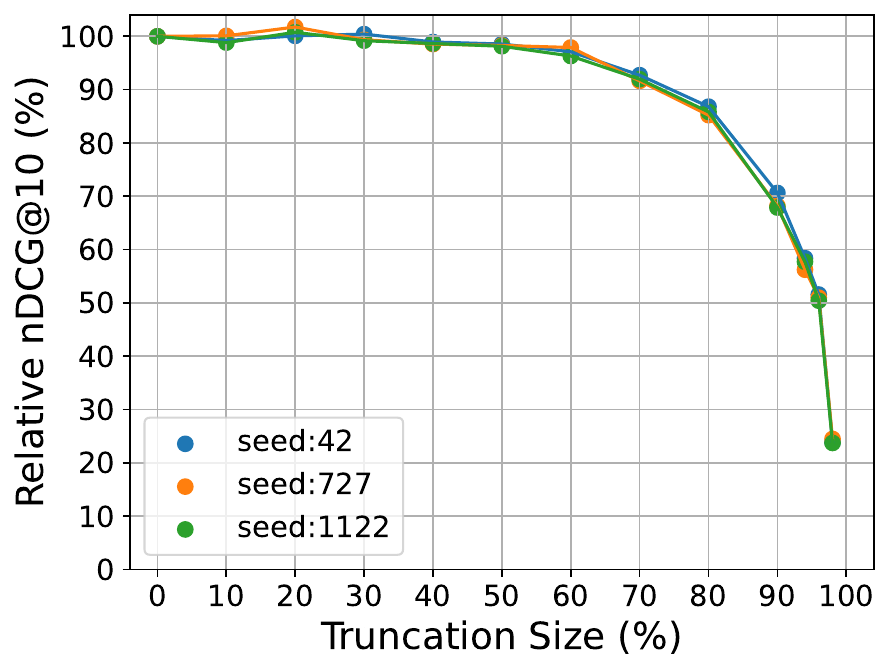}
    \caption{Relative performance of three differently-seeded models on NanoBEIR benchmark with different truncation sizes.}
    \label{fig:different_seeds}
\end{figure}
Here, we investigate whether the embeddings produced by all of the differently-seeded models during contrastive training achieve high relative performance by fine-tuning the T5's encoder with a contrastive learning objective as done in \S \ref{sec:anisotropy} with three different random seeds.
The result is shown in Fig. \ref{fig:different_seeds}.
All three model instances have almost identical shifts in relative performance with different degrees of truncations.
This result allows us to conclude that our observation, the embeddings' high robustness to the truncation operation, is present regardless of the randomness during contrastive training.

\subsection{LLMs and Truncated Representations}
\begin{table}[h]
    \centering
    \setlength{\tabcolsep}{15pt}
    \small
    \begin{tabular}{ccrrr}
    \toprule
    \textbf{Model} & \multicolumn{1}{c}{\textbf{Method}} & \multicolumn{1}{c}{\textbf{COPA F1 (Relative)}} & \multicolumn{1}{c}{\textbf{DROP F1 (Relative)}} & \multicolumn{1}{c}{\textbf{HellaSwag Acc (Relative)}} \\
    \midrule
    \multirow{3}{*}{Llama} & Full & 0.194 (1.000) & 0.194 (1.000) & 0.681 (1.000) \\
    \cmidrule{2-5}
     & First & 0.094 (0.487) & 0.094 (0.487) & 0.580 (\textbf{0.852})\\
     & Last & 0.038 (0.198) & 0.038 (0.198) & 0.586 (\textbf{0.861})  \\
    \midrule
    \multirow{3}{*}{Qwen} & Full & 0.003 (1.000) & 0.003 (1.000) & 0.718 (1.000) \\
    \cmidrule{2-5}
     & First & 0.001 (0.375) &  0.001 (0.375) & 0.709 (\textbf{0.988}) \\
     & Last & 0.001 (0.458) & 0.001 (0.458) & 0.709 (\textbf{0.987})\\
    \bottomrule
    \end{tabular}
    \caption{Performance on three benchmark datasets when the last hidden representations and the unembedding layer are reduced by half. Relative performance (scores in parenthesises) is bolded when it reaches 80\% of the original performance.}
    \label{tab:llm-gen-more-datasets}
\end{table}
Table \ref{tab:llm-gen-more-datasets} is a complementary table to Table \ref{tab:llm-gen}, showing the impact of representation truncation on the three remaining datasets from \S \ref{sec:k-removal}.

\subsection{Document Ranking by Truncated Embeddings}
\begin{figure}[h]
    \centering
    \includegraphics[width=0.8\textwidth]{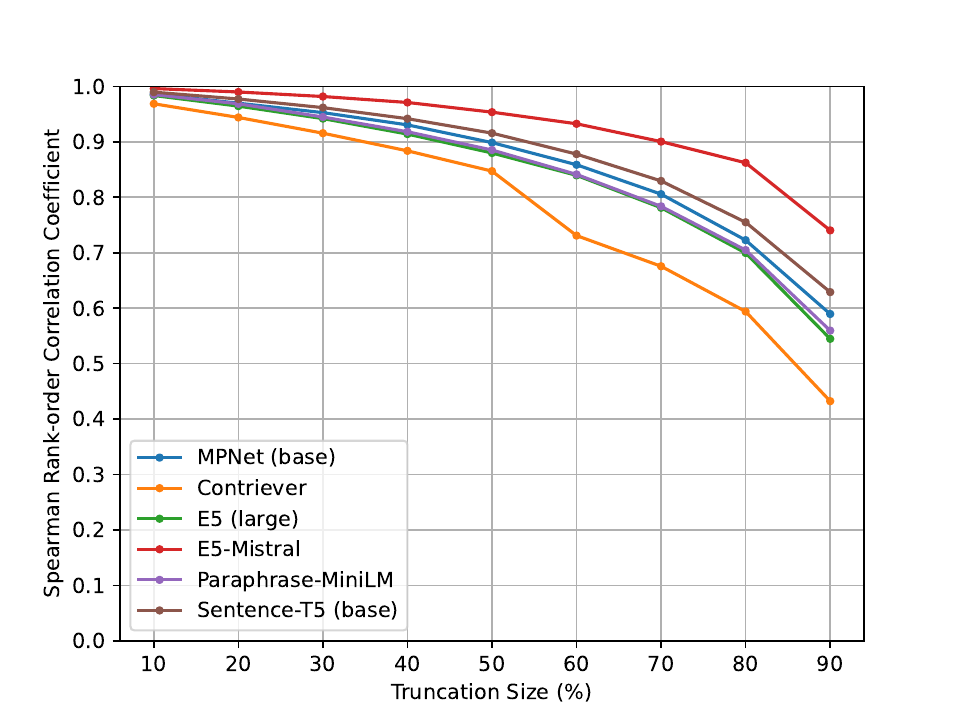}
    \caption{Spearman's rank-order correlation coefficient between the two rankings of documents for each query in NanoBEIR produced by full-sized embeddings and truncated embeddings.}
    \label{fig:rank_analysis_nanobeir}
\end{figure}
We compare two sets of document rankings.
The first set is a result of standard text embedding-based retrieval, where we use the original full-sized embeddings to compute similarity between queries and documents.
The second set is produced by using embeddings of which the last K\% of the dimensions are truncated.
We compute Spearman's rank-order correlation between the two and see how similar the rankings produced by truncated embeddings are to the original embeddings' rankings.
The resulting curve is shown in Fig. \ref{fig:rank_analysis_nanobeir}.
Similarly to our findings in \S \ref{sec:k-removal}, the behaviour of truncated embeddings is close to the original embeddings.

\newpage
\subsection{Truncated Embeddings' Performance Per Dataset}
\begin{figure*}[h]
    \centering
    \small
    \includegraphics[width=1.0\textwidth]{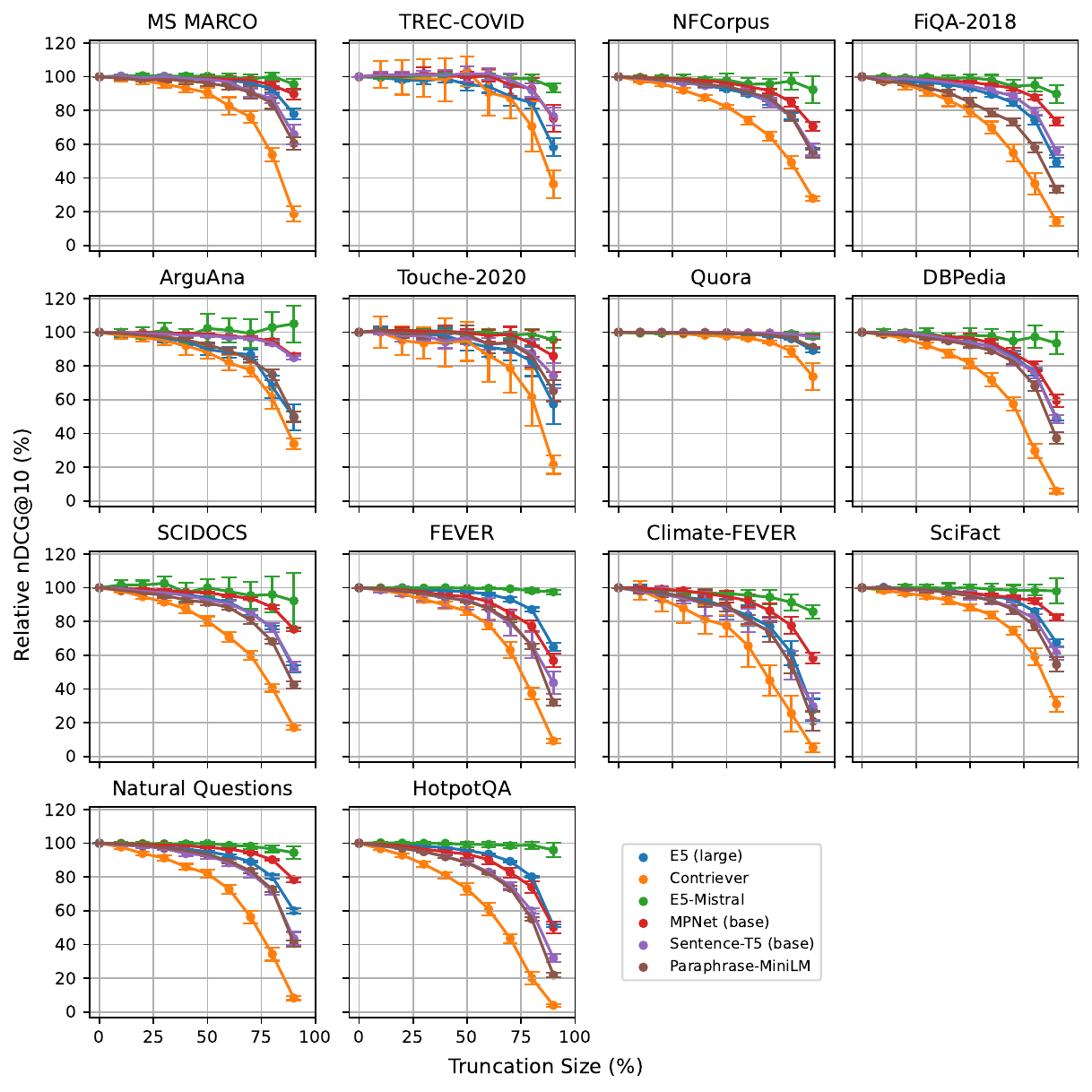}
    \caption{
    The relative performance achieved by randomly truncated embeddings per dataset from BEIR and NanoBEIR for E5-Mistral.
    }
    \label{fig:k-truncation-last-beir-datasets}
\end{figure*}
\newpage
\begin{figure*}[h]
    \centering
    \small
    \includegraphics[width=1.0\textwidth]{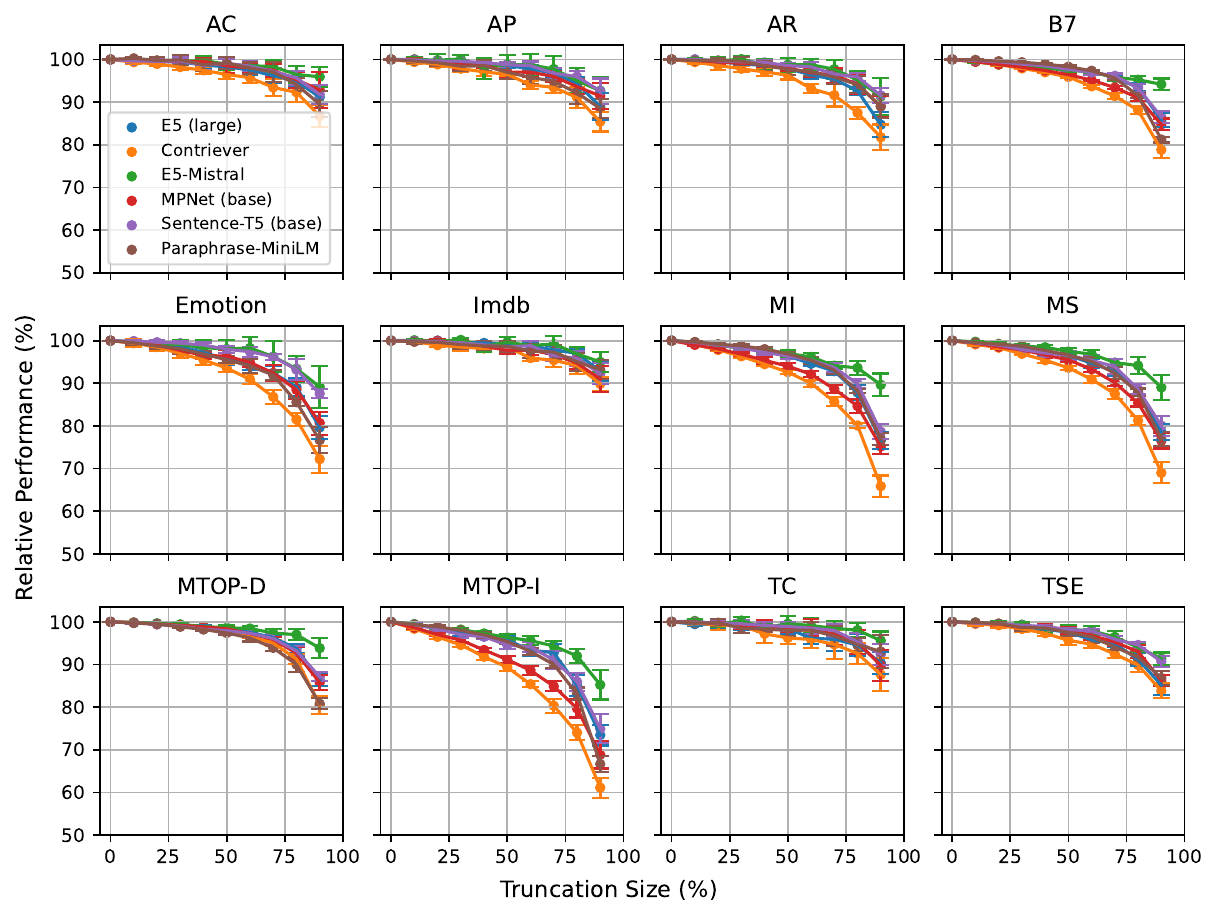}
    \caption{
    The relative performance achieved by randomly truncated embeddings per dataset from MTEB.
    }
    \label{fig:k-truncation-last-mteb-datasets}
\end{figure*}

\newpage
\subsection{Effective Use of Representation Space (Figures for BERT)}
\begin{figure*}[h]
    \centering
    \includegraphics[width=1\textwidth]{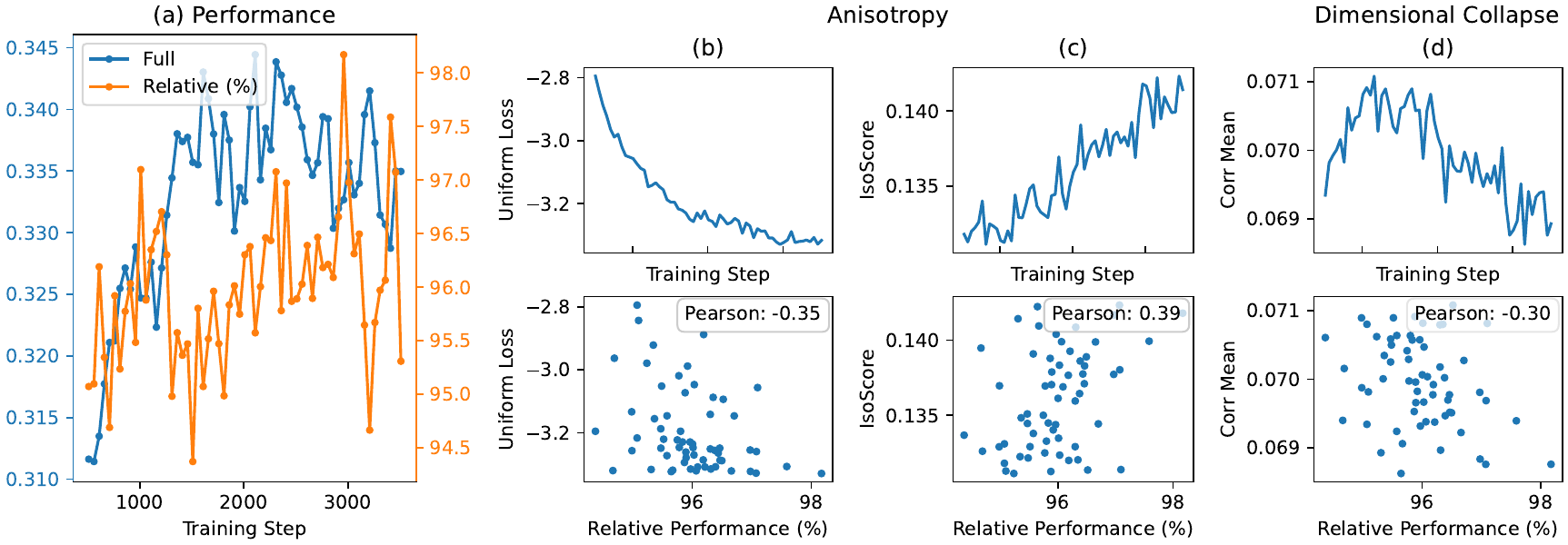}
    \caption{
    As a result of contrastive learning for BERT, downstream task performance increases (a: Full), and the use of embedding space measured through Uniform Loss ($\downarrow$) and IsoScore ($\uparrow$) for anisotropy (b, c: top) and Corr Mean ($\downarrow$) for dimensional collapse (d: top) also improves.
    However, the relative performance does not change over the training (a: Relative), therefore, there is no strong correlation between relative performance and representation quality measures (b, c, d: bottom).
    }
    \label{fig:existing-theories-bert}
\end{figure*}

\newpage
\subsection{Effect of Removing Only Degrading (or Improving) Dimensions}
\begin{figure}[h]
    \centering
    \small
    \includegraphics[width=1.0\columnwidth]{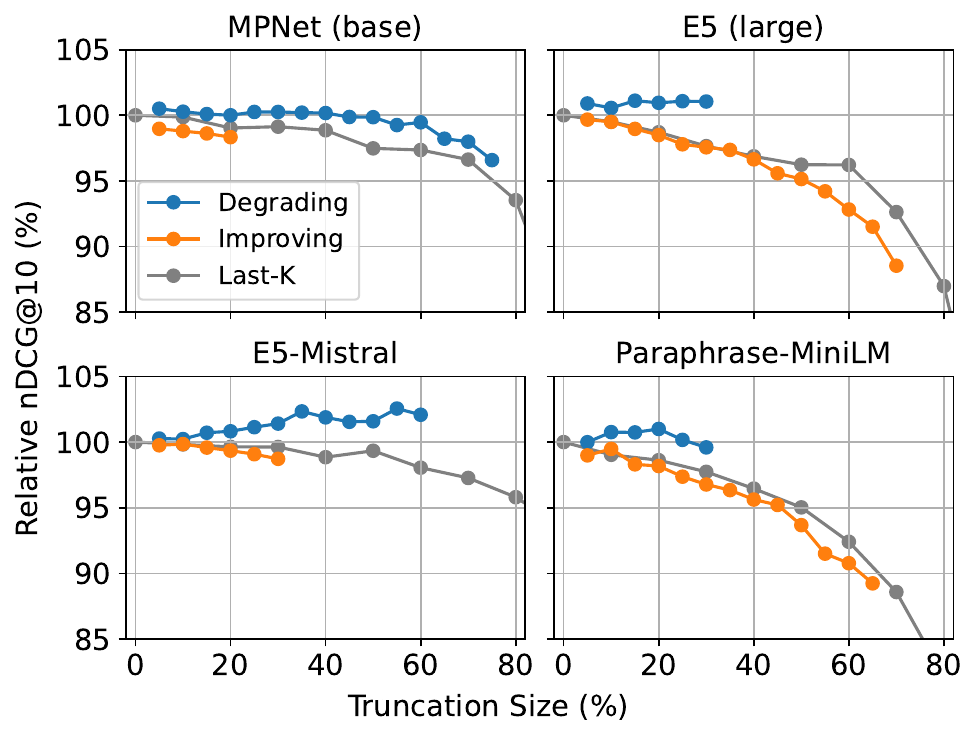}
    \caption{When only the degrading dimensions are removed (blue plot), the performance improves over the original embeddings first, and as more are removed, the performance starts to decay, however, more slowly than the last-k truncation. On the other hand, when only the improving dimensions are removed (orange plot), the performance decreases rapidly.}
    \label{fig:truncate-dd-id-only-others}
\end{figure}

\newpage
\subsection{E5-Mistral's Outlier Dimensions.}
\begin{figure}[h]
    \centering
    \includegraphics[width=0.8\linewidth]{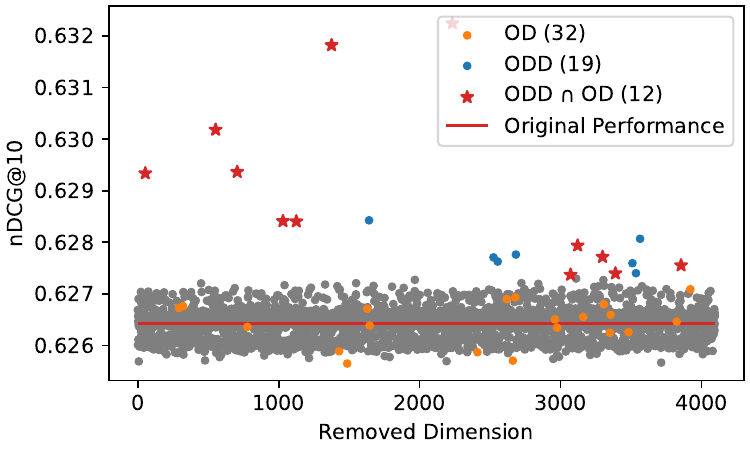}
    \caption{E5-Mistral's retrieval performance without one dimension. Outlier dimensions (ODs) take abnormal values within embeddings as defined in \S \ref{sec:outlier-dimensions}. Outlier degrading dimensions (ODDs) have negative impacts that are at least $3\sigma$ from the mean of performance. The numbers in parentheses indicate the count of each type of dimension.}
    \label{fig:outlier-highlighted-dd}
\end{figure}

\newpage
\subsection{Resources Used for Our Experiments}
\begin{table*}[h]
    \centering
    \small
    \setlength{\tabcolsep}{2pt}
    \begin{tabular}{crrc}
    \toprule
    \textbf{Name} & \textbf{Params} & \textbf{Emb Size} & \textbf{Licence} \\
    \midrule
    \href{https://huggingface.co/sentence-transformers/all-mpnet-base-v2}{MPNet (base)} & 109M & 768 & Apache license 2.0 \\
    \href{https://huggingface.co/facebook/contriever}{Contriever}~\citep{izacard_unsupervised_2022} & 110M & 768 & Attribution-NonCommercial 4.0 International \\
    \href{https://huggingface.co/intfloat/e5-large-v2}{E5 (large)}~\citep{wang_text_2022} & 335M & 1024 & MIT \\
    \href{https://huggingface.co/intfloat/e5-mistral-7b-instruct}{E5-Mistral}~\citep{wang_improving_2024} & 7B & 4096 & MIT \\
    \href{https://huggingface.co/sentence-transformers/paraphrase-MiniLM-L3-v2}{Paraphrase-MiniLM}~\citep{reimers_sentencebert_2019} & 17M  & 384 & Apache license 2.0 \\
    \href{https://huggingface.co/sentence-transformers/sentence-t5-base}{Sentence-T5 (base)}~\citep{ni_sentencet5_2022} & 110M & 768 & Apache license 2.0 \\
    \bottomrule
    \end{tabular}
    \caption{List of models used in our study.}
    \label{tab:model-list}
\end{table*}
\begin{table*}[h]
    \footnotesize
    \centering
    \setlength{\tabcolsep}{1.5pt}
    \begin{tabular}{llll}
      \toprule
      
       & \multicolumn{1}{c}{\textbf{Name}} & \multicolumn{1}{c}{\textbf{Domain}} &  \multicolumn{1}{c}{\textbf{Licence}} \\
      \midrule
      \multirow{14}{*}{\rotatebox[origin=c]{90}{\textbf{Retrieval}}}& \href{https://microsoft.github.io/msmarco/}{MS MARCO}~\citep{nguyen2016ms} & Misc. & MIT \\
      & \href{https://ir.nist.gov/covidSubmit/index.html}{TREC-COVID}~\citep{TREC-COVID} & Bio-Medical & Dataset License Agreement \\
      & \href{https://www.cl.uni-heidelberg.de/statnlpgroup/nfcorpus/}{NFCorpus}~\citep{boteva2016full} & Bio-Medical & N/A\\
      & \href{https://sites.google.com/view/fiqa/}{FiQA-2018}~\citep{fiqa} & Finance & N/A \\
      & \href{http://argumentation.bplaced.net/arguana/data}{ArguAna}~\citep{wachsmuth_retrieval_2018} & Misc. & CC BY 4.0 \\
      & \href{https://webis.de/events/touche-20/shared-task-1.html}{Touche-2020}~\citep{bondarenko2020overview} & Misc. & CC BY 4.0 \\
      & \href{https://www.quora.com/q/quoradata/First-Quora-Dataset-Release-Question-Pairs}{Quora}& Quora & N/A \\
      & \href{https://github.com/iai-group/DBpedia-Entity/}{DBPedia}~\citep{dbpedia} & Wikipedia & CC BY-SA 3.0 \\
      & \href{https://allenai.org/data/scidocs}{SCIDOCS}~\citep{cohan_specter_2020} & Scientific & GNU General Public License v3.0 \\
      & \href{http://fever.ai/}{FEVER}~\citep{thorne_fever_2018} & Wikipedia & CC BY-SA 3.0 l \\
      & \href{http://climatefever.ai/}{Climate-FEVER}~\citep{leippold2020climatefever} & Wikipedia & N/A \\
      & \href{https://github.com/allenai/scifact}{SciFact}~\citep{wadden_fact_2020} & Scientific & CC BY-NC 2.0 \\
      & \href{https://github.com/google-research-datasets/natural-questions}{Natural Questions}~\citep{kwiatkowski_natural_2019} & Scientific & CC BY-SA 3.0 \\
      & \href{https://hotpotqa.github.io/}{HotpotQA}~\citep{yang_hotpotqa_2018} & Scientific & CC BY-SA 4.0 \\
      \midrule

      \multirow{12}{*}{\rotatebox[origin=c]{90}{\textbf{Classification}}}& \href{https://huggingface.co/datasets/mteb/amazon_counterfactual}{AmazonCounterfactualClassification}~\citep{oneill_wish_2021} & Reviews, Written & CC-by-4.0 \\
      & \href{https://huggingface.co/datasets/mteb/amazon_polarity}{AmazonPolarityClassification}~\citep{10.1145/2507157.2507163} & Reviews, Written & Apache 2.0 \\
      & \href{https://huggingface.co/datasets/mteb/AmazonReviewsClassification}{AmazonReviewsClassification}~\citep{keung_multilingual_2020} & Reviews, Written & N/A \\
      & \href{https://huggingface.co/datasets/mteb/banking77}{Banking77Classification}~\citep{casanueva_efficient_2020} & Written & MIT \\
      & \href{https://huggingface.co/datasets/mteb/emotion}{EmotionClassification}~\citep{saravia_carer_2018} & Social, Written & N/A \\
      & \href{https://huggingface.co/datasets/mteb/imdb}{ImdbClassification}~\citep{maas_learning_2011} & Reviews, Written & N/A \\
      & \href{https://huggingface.co/datasets/mteb/amazon_massive_intent}{MassiveIntentClassification}~\citep{fitzgerald_massive_2023} & Spoken & Apache 2.0 \\
      & \href{https://huggingface.co/datasets/mteb/amazon_massive_scenario}{MassiveScenarioClassification}~\citep{fitzgerald_massive_2023} & Spoken & Apache 2.0 \\
      & \href{https://huggingface.co/datasets/mteb/mtop_domain}{MTOPDomainClassification}~\citep{li_mtop_2021} & Spoken & N/A \\
      & \href{https://huggingface.co/datasets/mteb/mtop_intent}{MTOPIntentClassification}~\citep{li_mtop_2021} & Spoken & N/A \\
      & \href{https://huggingface.co/datasets/mteb/toxic_conversations_50k}{ToxicConversationsClassification}~\citep{jigsaw-unintended-bias-in-toxicity-classification} & Social, Written & CC-by-4.0 \\
      & \href{https://huggingface.co/datasets/mteb/tweet_sentiment_extraction}{TweetSentimentExtractionClassification}~\citep{tweet-sentiment-extraction} & Social, Written & N/A \\

      \bottomrule
      
    \end{tabular}
    \caption{A list of datasets used in our evaluation.}
    \label{tab:datasets}
\end{table*}

\end{document}